\documentclass{article}

 \usepackage[main, final]{neurips_2026}

\usepackage[utf8]{inputenc} 
\usepackage[T1]{fontenc}    
\usepackage{hyperref}       
\usepackage{url}            
\usepackage{booktabs}       
\usepackage{amsfonts}       
\usepackage{nicefrac}       
\usepackage{microtype}      
\usepackage{xcolor}         

\usepackage{graphicx}
\usepackage{subcaption}

\usepackage{amsmath}
\usepackage{amssymb}
\usepackage{mathtools}
\usepackage{amsthm}
\usepackage{bm}
\usepackage{float}
\usepackage{wrapfig}

\newcommand{\R}{\mathbb{R}}

\newcommand{\g}{\bm{g}}
\newcommand{\p}{\bm{p}}

\newcommand{\x}{\bm{x}}
\newcommand{\W}{\bm{W}}
\newcommand{\s}{\bm{s}}

\renewcommand{\L}{\mathcal{L}}
\renewcommand{\a}{\mathbf{a}}
\renewcommand{\b}{\mathbf{b}}

\newcommand{\muP}{\mu\text{P}}

\usepackage[capitalize,noabbrev]{cleveref}
\usepackage{multirow}
\theoremstyle{plain}
\newtheorem{theorem}{Theorem}[section]
\newtheorem{proposition}[theorem]{Proposition}

\theoremstyle{definition}
\newtheorem{definition}[theorem]{Definition}
\newtheorem{assumption}[theorem]{Assumption}
\newtheorem{fact}[theorem]{Fact}
\newtheorem{lemma}[theorem]{Lemma}
\theoremstyle{remark}
\newtheorem{remark}[theorem]{Remark}

\title{Towards Hyperparameter Transfer for Differentially Private Optimization}

\author{%
  Tianze Wang \\
  Department of Statistics\\
  Rutgers University\\
  \texttt{tianze.wang@rutgers.edu} \\ 
  \And
  Zhiqi Bu\footnotemark[1] \\
  Independent Researcher \\
  \texttt{woodyx218@gmail.com} \\
  \And
  Linjun Zhang\thanks{Co-corresponding authors.} \\
  Department of Statistics\\
  Rutgers University\\
  \texttt{zlj11112222@gmail.com} \\ 
}
\begin{document}

\maketitle

\begin{abstract}
Scaling laws and hyperparameter tuning remain central challenges in training large-scale models, particularly under differential privacy (DP), where optimization is further complicated by noise injection and per-sample gradient clipping. While Maximal Update Parametrization ($\mu$P) allows for zero-shot hyperparameter transfer in standard training, we show that it fails under DP constraints due to the distinct spectral properties of high-dimensional noise versus low-rank gradients. In this work, we introduce DP-$\mu$P, a theoretically grounded framework that extends $\mu$P to the privacy-preserving setting. By analyzing the interaction between gradient clipping, noise injection, and feature learning, we derive a Signal-to-Noise Ratio (SNR) condition that governs layer-wise learning rate scaling. Combining this with our proposed spectral scaling formulation, we provide specific scaling rules for DP-SGD, DP-Adam, and DP-Muon optimizers. Empirically, we demonstrate that DP-$\mu$P enables effective zero-shot transfer of learning rates across model widths on diverse architectures (MLP, ViT, GPT-2), matching or exceeding the performance of individually tuned baselines while significantly reducing computational costs.
\end{abstract}

\section{Introduction}
\label{sec:intro}
The success of large-scale models such as GPT \cite{radford2019language, brown2020languagemodelsfewshotlearners}, LLaMA \cite{touvron2023llama}, Qwen \cite{yang2024qwen2}, and DeepSeek \cite{liu2024deepseek}, trained with billions of parameters and trillions of tokens, has underscored the importance of effective hyperparameter tuning \cite{hu2022lora, houlsby2019parameter, karimi2021compacter}. Empirically, model performance is sensitive to choices like learning rate, momentum, and weight decay, necessitating fine-grained and computationally intensive tuning. However, the resources used for tuning hyperparameters grow exponentially with model size under grid or random search,
posing a development for scalable model training \cite{raffel2023exploringlimitstransferlearning, chowdhery2023palm}.

Maximal Update Parametrization ($\muP$) \cite{yang2020feature} has emerged as a compelling solution to hyperparameter tuning in large-scale models. By reparameterizing neural networks to preserve update magnitudes across widths \cite{yang2021tensor, yang2022tensor}, $\muP$ enables zero-shot transfer of hyperparameters, such as learning rate, from small models to significantly larger ones without performance degradation. $\muP$ achieves this by ensuring consistent layer-wise updates that stabilize activations and weights across both training iterations and model layers. 
In contrast to empirical scaling laws, which can be inaccurate, model-specific, or computationally expensive, $\muP$ offers a theoretically grounded, architecture-agnostic framework for scalable training and robust feature learning \cite{yang2023spectral}.

In this work, we investigate $\mu$P in differentially private (DP) optimization \citep{abadi2016deep} to enable hyperparameter (learning rate) transfer in privacy-preserving deep learning. To put this into perspective, a well-known empirical tradeoff exists between model utility and privacy risk. On the one hand, larger models typically achieve better performance in tasks such as image classification and language modeling \cite{he2016deep, brown2020languagemodelsfewshotlearners}. On the other hand, these models are also more prone to memorizing training data, which can be extracted through membership inference attacks \cite{shokri2017membershipinferenceattacksmachine} or prompt-based data extraction techniques \cite{carlini2021extractingtrainingdatalarge}. To protect data privacy while retaining model utility, DP can be equipped with any optimizer. Moreover, DP has proven effective in settings such as continual learning and fine-tuning~\cite{hassanpour2022differential, bu2020deep, yu2022differentiallyprivatefinetuninglanguage}, making it a practical and increasingly adopted approach for privacy-preserving machine learning.

Specifically, we consider DP optimizers equipped with layer-wise clipping \cite{bu2023accuracyefficiencygroupwiseclipping, bu2023convergencecalibrationdeeplearning, he2022exploringlimitsdifferentiallyprivate, mcmahan2018learningdifferentiallyprivaterecurrent}. Our objective is to quantify the layer-wise stabilization required to ensure robust learning rate transfer in DP optimization. {In particular, we focus on the noise-dominated regime, where the spectral norm of the injected privacy noise overwhelms the gradient signal. This is an inevitable scenario that naturally arises in large models due to the high dimensionality of the noise matrix \cite{tramèr2021differentiallyprivatelearningneeds}, and is a standard setting commonly analyzed in DP deep learning \cite{vogels2020powersgdpracticallowrankgradient, yu2021letprivacyoverbillutility, li2022largelanguagemodelsstrong}. However, we emphasize that noisy learning frameworks (including both DP training and zero-order optimization \cite{malladi2024finetuninglanguagemodelsjust}) can still achieve meaningful utility even when noise dominates in magnitude.} We mark our contributions as follows:
\begin{enumerate}
    \item \textbf{[Spectral condition for DP]} We characterize the spectral properties of the noise-dominated regime using random matrix theory: the full-rank nature of high-dimensional Gaussian noise, in contrast to the stable low-rank structure of the gradient signal, induces fundamentally different spectral dynamics that dictate the optimal scaling rules.

    \item \textbf{[Signal-to-noise condition for DP]} We derive a signal-to-noise ratio (SNR) condition under the $\mu$P framework, which, together with the spectral condition, enables learning rate transfer across model sizes. We demonstrate this for three optimizers: DP-SGD, DP-Adam, and DP-Muon, establishing novel scaling rules for both clipping-only and DP settings, summarized in \Cref{tab:lr and init muP}.

    \item \textbf{[Zero-shot hyperparameter transfer]} We present a comprehensive empirical study, including coordinate checks to support our scaling method. We demonstrate that our method effectively transfers optimal learning rates from small to large models, matching or exceeding the performance of individually tuned baselines.
\end{enumerate}

\begin{figure*}[!t]
\centering
\includegraphics[width=0.95\linewidth]{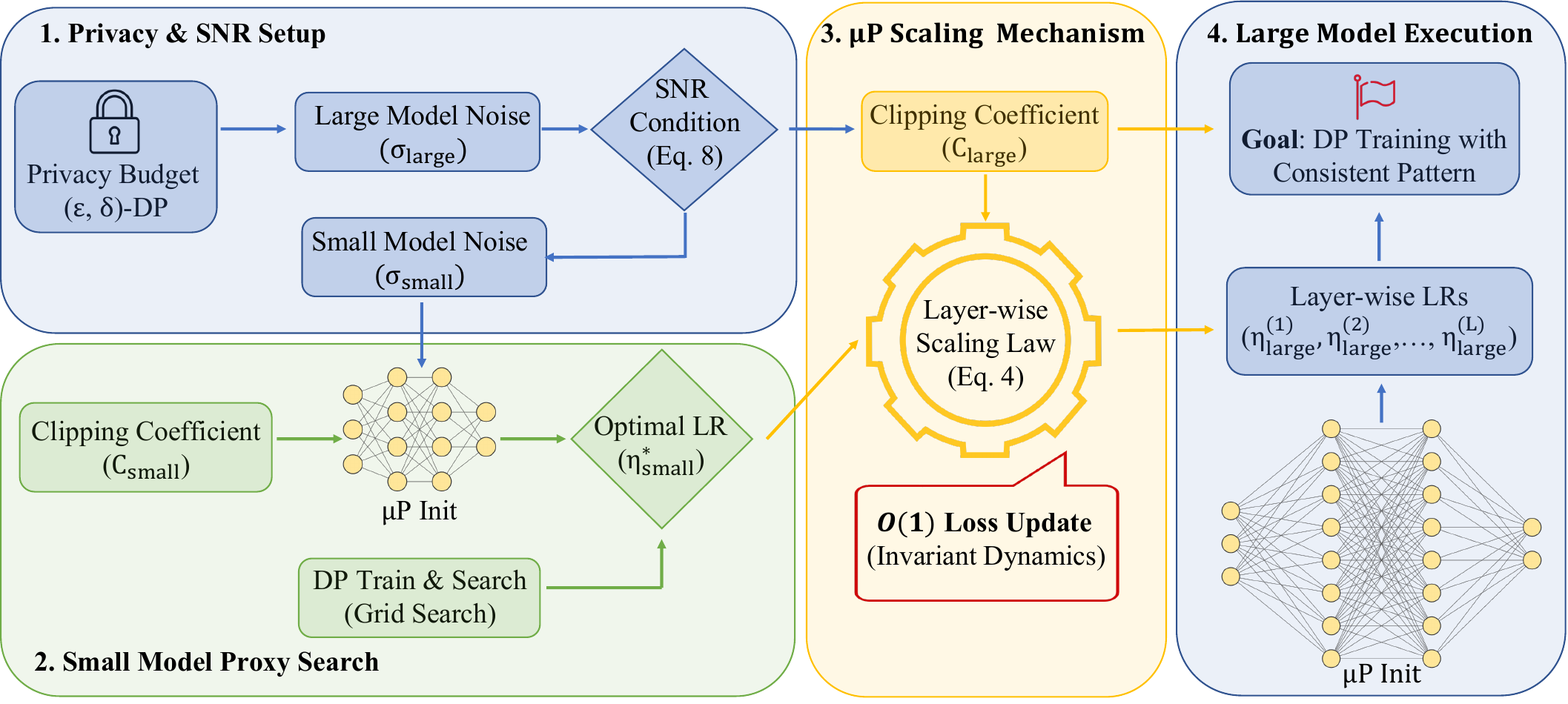}
\caption{\textbf{DP-$\mu$P Training Workflow.} We employ $\mu$P initialization to stabilize the initial state and align the Signal-to-Noise Ratio (SNR) to derive the appropriate noise parameters for the small proxy model. After identifying the optimal base learning rate on the proxy, we apply layer-wise scaling to the large model to ensure $\mathcal{O}(1)$ loss updates, resulting in invariant training dynamics across model scales. A numerical example is provided in Appendix~\ref{sec:derivation_example}.}
\label{fig:flowchart}
\end{figure*}

\begin{table}
  \caption{Summary of layer-wise learning rate (lr) scaling rules for $\mu$P under different privacy settings. Note that for DP-SGD, the scaling explicitly accounts for the noise-to-batch ratio.}
  \label{tab:lr and init muP}
  \centering
  \begin{tabular}{l|c|c|c}
    \toprule
    Optimizer         & NonDP lr scaling \cite{yang2022tensor}& Clipping-only lr scaling & DP lr scaling\\
    \midrule
    SGD  & $\frac{n_{l}}{n_{l-1}}$ 
         & $\sqrt{\frac{n_l}{n_{l-1}}}$ 
         & $\frac{\sqrt{n_l / n_{l-1}}}{\frac{\sigma_{\text{DP}}}{B}(\sqrt{n_l}+\sqrt{n_{l-1}})}$  \\
    Adam & $\frac{1}{n_{l-1}}$     
         & $\frac{1}{n_{l-1}}$          
         & $\frac{\sqrt{n_l / n_{l-1}}}{\sqrt{n_l}+\sqrt{n_{l-1}}}$\\
     Muon & $\sqrt{\frac{n_l}{n_{l-1}}}$ 
         & $\sqrt{\frac{n_l}{n_{l-1}}}$ 
         & $\sqrt{\frac{n_l}{n_{l-1}}}$ \\ 
    \bottomrule
  \end{tabular}
\end{table}

\section{Preliminary}
\subsection{Notations}
We use the big-$\Theta$ notation to describe asymptotic scaling, where $f(n) = \Theta(g(n))$ if there exist positive constants $c, C$ such that
\[
    c \cdot g(n) \le f(n) \le C \cdot g(n),\; \text{for all sufficiently large~} n.
\]
We denote the Frobenius norm by $\|\W\|_F := \sqrt{\sum_{i,j} \W_{ij}^2}$, and the spectral norm (or operator norm) by $\|\W\|_* := \max_{\mathbf{x} \neq 0} {\|\W\mathbf{x}\|_F}/{\|\mathbf{x}\|_F}$. {Finally, we use $\asymp$ to indicate asymptotic equivalence up to constant factors.}

\subsection{Deep Learning and Gradient Descent}
Consider a neural network composed of $L$ layers, where each layer transforms the output of the previous one. A simplified forward pass through the network can be represented layer-wise as follows:
\begin{equation} 
\label{eq:forward}
\a_{l} = \a_{l-1} \W_l,\quad \frac{\partial \L}{\partial \W_l}=\frac{\partial \a_l}{\partial \W_l}^\top\frac{\partial \L}{\partial \a_l}=\a_{l-1}^\top\frac{\partial \L}{\partial \a_l}
\end{equation}
where $l$ denotes the layer index. Here, $\a_l\in\R^{B\times n_{l}}$ represents the activation at layer $l$ for a mini-batch of size $B$, with $\a_0$ being the input data, and $\W_l\in\R^{n_{l-1}\times n_{l}}$ is the weight matrix of the $l$-th layer. Following the convention in $\muP$ literature \cite{yang2022tensor,yang2023spectral}, we omit bias terms and nonlinear activation functions (e.g., ReLU or GELU) between layers for notational simplicity. We highlight that our analysis also works without such simplification and explain it in Appendix~\ref{sec:bias and nonlinearities}.

Training neural networks typically involves iterative updates using gradient-based optimization algorithms (e.g. SGD, Adam and Muon). At each iteration $t$, the weights are updated according to:
\begin{equation} \label{eq:grad descent}
\W_{l,t+1}=  \W_{l,t}+\Delta \W_{l,t} :=  \W_{l,t}-\eta_l \g_{l,t}
\end{equation}
where $\eta_l$ is the learning rate associated with layer $l$, and $\g_{l,t}=g(\nabla_{\W_{l,t}} \L)$ is the pre-conditioned gradient of the mini-batch loss function $\L$. The function $g(\cdot)$ defines the optimizer-specific transformation applied to the raw gradient. For example, $g(\bm{x}) = \bm{x}$ corresponds to standard SGD; $g(\bm{x})=\text{sign}(\bm{x})$ corresponds to SignSGD or the gradient update in Adam-like optimizers; $g(\bm{x})=\bm{U^\top V}$ (with $\bm{x}=\bm{U^\top SV}$ being the singular value decomposition) corresponds to Muon and Shampoo.

\subsection{Maximal Update Parametrization ($\muP$)}
To ensure feature learning, $\muP$ stabilizes each element of $\a_l$ at $\Theta(1)$ in two dimensions: layer-wise and iteration-wise. Mathematically, $\muP$ normalizes the weights to satisfy the following scales:
\begin{equation}
\begin{aligned}
& \overbrace{\|\a_{l}\|_2=\Theta(\sqrt{n_l}) \Leftarrow \|\W_{l,t}\|_*=\Theta(\sqrt{n_{l}/n_{l-1}})}^{\substack{\text{Desideratum 1 in \cite{yang2023spectral}}~\text{(Feature learning)}}} \\
& \underbrace{\|\W_{l,t}\|_*=\Theta(\sqrt{n_{l}/n_{l-1}}) \Leftarrow \|\Delta\W_{l,t}\|_*=\Theta(\sqrt{n_{l}/n_{l-1}})}_{\substack{\text{Condition 1 in \cite{yang2023spectral}}~ \text{(Spectral scaling)}}}
\label{eq: conditions}
\end{aligned}
\end{equation}

Note that \eqref{eq: conditions} can be derived via mathematical induction (detailed in Appendix~\ref{app_lem:layerwise lr}), provided the following specifications are met:
\begin{enumerate}
    \item weight initialization with specific variance (known as the spectral parametrization in \cite{yang2023spectral}), so that at $t=0$, $\|\W_{l,0}\|_*=\Theta(\sqrt{n_l/n_{l-1}})$.
    \item layer-wise learning rate $\eta$, so that $\|\Delta\W_{l,t}\|_*=\Theta(\sqrt{n_{l}/n_{l-1}})$ for any $\g_{l,t}$. 
\end{enumerate}

Crucially, it has been established that preserving this feature learning regime is fundamental to enabling hyperparameter transfer across model widths \cite{yang2022tensor,yang2023spectral}.
{For residual connections in transformers, since $\mu$P meticulously scales the residual outputs to ensure the activation norms remain stable, the framework extends to residual architectures without loss of generality.} Consequently, to achieve this transferability in the differentially private setting, we determine the \textbf{layer-wise learning rate} specifically to enforce these spectral scaling laws. This requirement leads to the condition formalized in the following lemma.
\begin{lemma}[Spectral condition via layer-wise learning rate \cite{yang2023spectral}]
\label{lem:layerwise lr}
For $\g_l$ from any optimizer, the spectral scaling condition in \eqref{eq: conditions} is equivalent to setting
\begin{equation}
    \eta_l=\|\Delta\W_l\|_*/\|\g_l\|_*=
\Theta(\sqrt{n_l/n_{l-1}}/\|\g_l\|_*)
\end{equation}
\end{lemma}

We can leverage Lemma~\ref{lem:layerwise lr} to derive \Cref{tab:lr and init muP} by analyzing $\|\g_l\|_*$ under different optimziers. To see this, we elaborate in \Cref{sec:main} for both DP and non-DP optimizers, with a special focus on the influence of DP operations (i.e. per-sample gradient clipping and Gaussian noising).

\subsection{Differentially Private Optimization}
Recall the formal definition of $(\epsilon, \delta)$-DP.
\begin{definition}
    \cite{dwork2006our} A randomized algorithm $\mathcal{M}$ is $(\epsilon, \delta)$-differentially private if, for any two neighboring datasets $D$ and $D^\prime$ that differ in exactly one datapoint, and for all $\mathcal{S} \subset \text{Range}(\mathcal{M})$, it holds that:
    \begin{equation}
        \Pr [\mathcal{M}(D) \in \mathcal{S}] \leq e^\epsilon \Pr [\mathcal{M}(D^\prime) \in \mathcal{S}] + \delta.
    \end{equation}
\end{definition}

Differentially private (DP) optimization applies any optimizer with the \textbf{privatized gradient}:
$$p(\nabla_{\W} \L)=\frac{1}{B}\sum_{i=1}^B c_i\nabla_{\W}\L_i+\frac{\sigma_\text{DP}}{B} N(0,I),$$
hence transforming SGD to DP-SGD, as well as AdamW \cite{KingmaB14,loshchilovdecoupled}, Muon \cite{jordan2024muon}, GeN \cite{bu2024gradient,liutowards}, etc. Here $\nabla_{\W}\L_i$ is the per-sample gradient, $c_i$ is a constant to perform gradient clipping ( e.g. $c_i=\frac{1}{\|\nabla_{\W}\L_i\|_F}$), $B$ is the batch size and $\sigma_\text{DP}$ is the noise level.

We study the layer-wise automatic clipping \cite{bu2023accuracy,yang2022normalized}, for the $l$-th layer,
\begin{align}
\p_l:=p(\nabla_{\W_l} \L)=\frac{1}{B}\sum_i \frac{\nabla_{\W_l}\L_i}{\sqrt{D}\|\nabla_{\W_l}\L_i\|_F}+\frac{\sigma_\text{DP}}{B}N(0,I_l)
\end{align}
where $D$ is the number of layers. We focus on this type of method for two reasons: (I) automatic clipping avoids hyperparameter tuning for the per-sample clipping threshold, and (II) we choose layer-wise clipping \cite{mcmahan2018learning} because $\mu$P requires a layer-wise analysis, and because layer-wise clipping is the most efficient and scalable clipping style in distributed training \cite{bu2023zero, bu2023accuracy}. 
To summarize, the pre-conditioned and privatized gradient is $\g_l=g(\p_l)$ for any optimizer $g$, which has the same level of DP guarantee by the post-processing property of DP.

\section{$\mu$P for DP optimizers}

\label{sec:main}
In this section, we analyze the spectral scaling properties of DP optimizers to address the unique challenges introduced by the privacy mechanism. We demonstrate that the interaction between clipping and noise invalidates standard $\mu$P transfer rules, necessitating a rigorous derivation of clipping thresholds and noise levels that satisfy a specific SNR constraint. Ultimately, we establish that learning rate transfer under DP requires two conditions to hold simultaneously: the $\mu$P spectral condition, which keeps the update at the feature-learning scale, and the layer-wise SNR condition, which keeps the share of signal in that update consistent across widths. {The SNR condition is necessary, and together with the spectral condition it is sufficient (Proposition~\ref{prop:snr_condition}).} Our analysis relies on the following fact and assumption.
\begin{fact}[Norm equivalence for per-sample gradient]
\label{assum:2}
$\|\nabla_{\W_l}\L_i\|_*= \|\nabla_{\W_l}\L_i\|_F$. Since the gradient is computed on a per-sample basis (equivalent to $B=1$), it is a rank-1 matrix.
\end{fact}

\begin{assumption}
\label{assum:alignment}
$\left\|\frac{1}{B}\sum_{i=1}^B g(\p_i)\right\|_* = \Theta\left(\frac{1}{B}\sum_{i=1}^B\|g(\p_i)\|_*\right)$. 
This geometric alignment assumption for gradients within a mini-batch is widely used for analyzing large-batch training \citep{yin2018gradientdiversitykeyingredient, vaswani2021painlessstochasticgradientinterpolation}. {We emphasize that this assumption is practical, with empirical validation provided in Appendix \ref{sec:empirical_alignment}.} 
\end{assumption}

\subsection{Standard Optimizers with Per-Sample Gradient Clipping}
\begin{figure}[ht]
    \centering
    \begin{subfigure}[b]{0.48\textwidth}
        \centering
        \includegraphics[width=\linewidth]{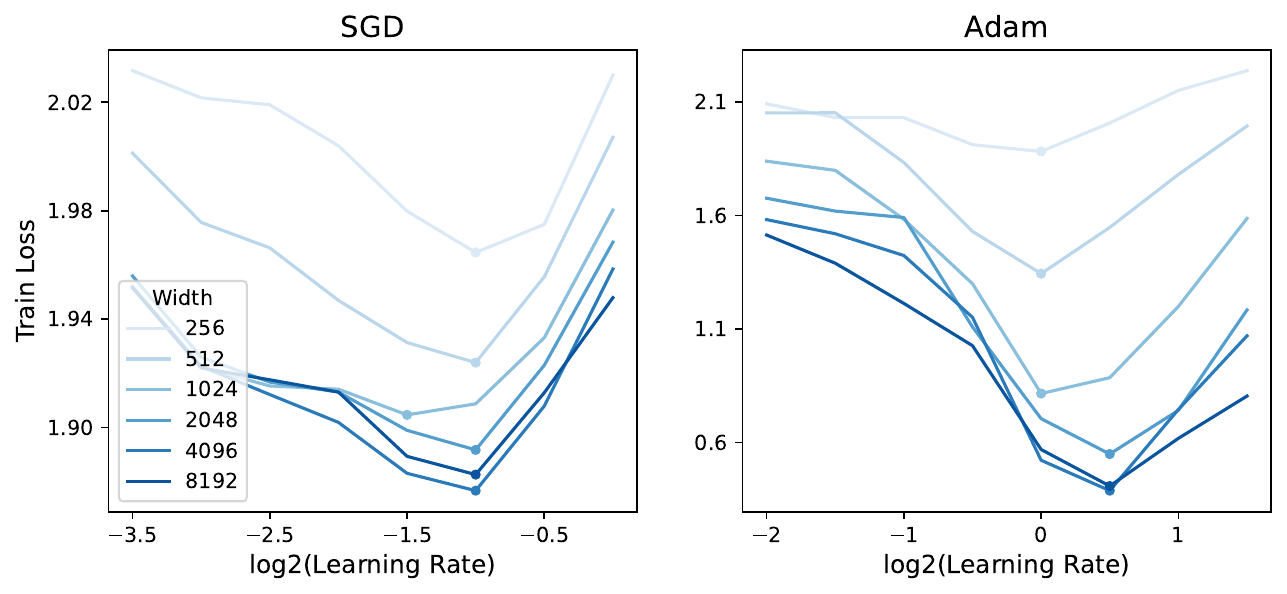}
        \caption{Clipping Only (Success)}
        \label{fig:mlp-clipping-only}
    \end{subfigure}
    \begin{subfigure}[b]{0.48\textwidth}
        \centering
        \includegraphics[width=\linewidth]{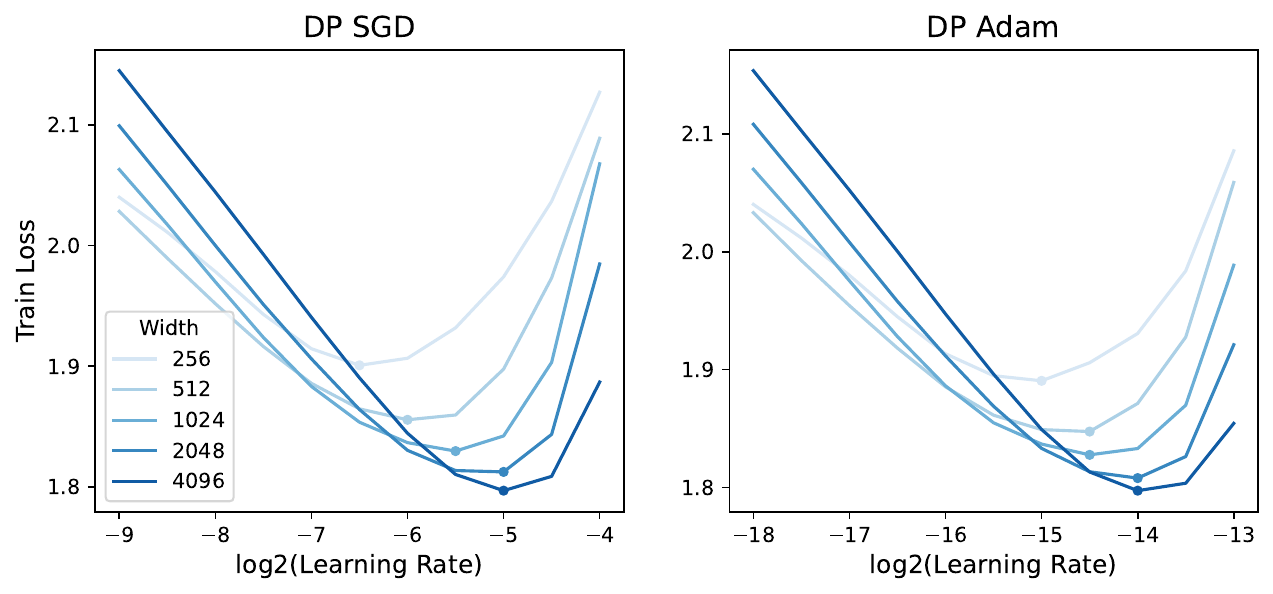}
        \caption{Noisy Norm Scaling (Failure)}
        \label{fig:mlp-failure-no-snr}
    \end{subfigure}

    \caption{\textbf{Comparison of hyperparameter transfer scaling rules ($\muP$) on a 5-layer MLP (CIFAR-10).} 
    {(a) Under the clipping-only regime (no noise), loss curves align across widths, validating $\muP$. 
    (b) Scaling by the spectral norm of the noisy update fails to align curves, demonstrating the failure of standard $\muP$ under differential privacy.}}
    \label{fig:mlp-comparison}
\end{figure}

Begin with analyzing DP optimizers without noise, which effectively reduces to standard optimizers with per-sample gradient clipping, we isolate the scaling behavior induced by clipping.
\begin{proposition}
The spectral norms of the clipped gradient updates $\g(\p_l)$ under different optimizers satisfy:
\begin{equation}
\label{eq:grad_norms}
\|g(\p_l)\|_* =
\begin{cases}
    \Theta(1), & \text{SGD / Muon}, \\
    \Theta\left(\sqrt{n_{l} n_{l-1}}\right), & \text{Adam}.
\end{cases}
\end{equation}
\end{proposition}
Substituting these estimates into Lemma~\ref{lem:layerwise lr} yields the scaling rules listed in the second column of \Cref{tab:lr and init muP}. We empirically validate these derivations in \Cref{fig:mlp-clipping-only}, which demonstrates successful learning rate transfer across model widths.


For input layer, the feature norm follows $\|\a_1\| = \Theta(\|\W_1\|_* \cdot \|\bm{x} \|)$. This behavior depends on the data norm $\|\bm{x}\|$, distinguishing from hidden layers where scaling is governed solely by the width of the preceding layer. We provide a detailed derivation of this boundary condition in Appendix~\ref{sec:input_layer_scaling}.

\begin{remark}
For non-noisy optimizers, the spectral condition in Lemma~\ref{lem:layerwise lr} is necessary and sufficient for transfer. Fundamentally, $\mu$P operates by scaling gradient updates to preserve feature learning as width increases. Gradient clipping operates as a scaling adjustment and thus remains within the $\mu$P framework.
Therefore, standard $\mu$P guarantees apply directly provided feature learning holds.
\end{remark}

\subsection{DP Optimizers with Per-Sample Gradient Clipping and Noise}
\label{sec:proof in main}
However, the spectral condition in Lemma~\ref{lem:layerwise lr} alone is insufficient for DP optimizers. In the privacy-preserving setting, the total update norm is often dominated by the injected noise rather than the signal. Scaling rules derived solely from the total spectral norm fail to preserve the feature learning dynamics, causing loss curves diverge significantly across widths (\Cref{fig:mlp-failure-no-snr}). We formally characterize this insufficiency in the following Lemma~\ref{lem:noise_insufficiency}.


\begin{lemma}[Spectral condition is insufficient for DP optimizers]
    \label{lem:noise_insufficiency}
    In the noise-dominated regime of SGD, the signal update vanishes as width increases. 
\end{lemma}

\begin{proof}
    Let the signal $\s_l$ and noise $\bm{\bm{\xi}}_l$ be defined as:
    \[
        \s_l = \frac{1}{B}\sum_{i=1}^B \frac{\nabla_{\W_l}\mathcal{L}_i}{\sqrt{D}\,\lVert \nabla_{\W_l}\mathcal{L}_i\rVert_F},
        \quad
        \bm{\xi}_l \sim \frac{\sigma_{\mathrm{DP}}}{B}\mathcal{N}(0, I_{n_l \times n_{l-1}})
    \]
    Consider the noise-dominated regime, where the spectral norm of the update is dominated by the random matrix contribution. Using standard random matrix theory results, we have:
    \[
        \lVert \s_l+\bm{\xi}_l\rVert_* = \Theta \left(\lVert \bm{\bm{\xi}}_l\rVert_*\right) = \Theta\left( \frac{\sigma_{\mathrm{DP}}}{B}\left(\sqrt{n_l} + \sqrt{n_{l-1}}\right)\right).
    \]
    {The first equality holds as $\|\s_l\| = \Theta(1)$ due to automatic clipping, which remains non-vanishing even when iterates near stationary.} After applying the scaling suggested in Lemma~\ref{lem:layerwise lr}, the effective update to the weight matrix $W_l$ with respect to the signal scales as:
    \[
        \Vert \eta_l \s_l\Vert_* = \Theta\left(\sqrt{\frac{n_l}{n_{l-1}}}\frac{ \Vert \s_l \Vert_*}{\Vert\bm{\bm{\xi}}_l\rVert_*}\right) = \Theta\left(\frac{\sqrt{n_l/n_{l-1}}}{\sqrt{n_l} + \sqrt{n_{l-1}}}\right).
    \]
    This scales as $\Theta(1/\sqrt{n})$, which implies that the signal contribution vanishes as the width increases, thereby violating the feature learning scales required for transfer.
\end{proof}

Next, we introduce Proposition~\ref{prop:snr_condition}, which establishes a layer-wise SNR condition for hyperparameter transfer under DP. {We show that this condition is necessary for transfer, and that it is also sufficient when combined with the layer-wise learning rate of Lemma~\ref{lem:layerwise lr}.} To satisfy this condition at every layer, we replace the global clipping coefficient with a layer-dependent one, parametrized by $D_l$.

\begin{proposition}[SNR condition is necessary and sufficient for DP optimizers]
    \label{prop:snr_condition}
    We define the spectral noise-to-signal ratio $\mathcal{C}(\cdot)$ for the $l$-th layer as a function of the noise multiplier $\sigma$, clipping coefficient $1/\sqrt{D_l}$, and layer widths:
    \[
        \frac{\lVert \bm{\bm{\xi}}_l \rVert_*}{\lVert \s_l \rVert_*} \asymp \frac{\sigma\sqrt{D_l}}{B}\left(\sqrt{n_l} + \sqrt{n_{l-1}}\right):= \mathcal{C}(\sigma, D_l, n_l,n_{l-1}).
    \]
    Let $c_l^{\mathrm{target}}$ denote this value calculated using the target model's configuration ($\sigma^{\mathrm{target}}, D_l^{\mathrm{target}}, n^{\mathrm{target}}$), and $c_l^{\mathrm{proxy}}$ denote the corresponding value for the proxy model ($\sigma^{\mathrm{proxy}}, D_l^{\mathrm{proxy}}, n^{\mathrm{proxy}}$). {In the noise-dominated regime, hyperparameters tuned on the proxy transfer to the target if and only if}
    \[
        c_l^{\mathrm{proxy}} = c_l^{\mathrm{target}}, \quad \forall l \in \{1, \dots, L\}.
    \]
    This equality implies that the effective noise intensity relative to the signal is preserved layer-wise, independent of the model scale.
\end{proposition}

\begin{proof}
    {We first show sufficiency:} under the SNR condition defined in Proposition~\ref{prop:snr_condition}, the effective signal contribution to the weight update remains independent of the variable width $n_l$, thus enabling hyperparameter transfer.

    Consider the normalized update step used in $\mu$P within the noise-dominated regime. The update direction is normalized by the spectral norm of the total gradient estimate:
    \[
        \Delta \W_{\text{norm}} = \frac{\s_l + \bm{\bm{\xi}}_l}{\lVert \s_l + \bm{\bm{\xi}}_l \rVert_*} \asymp \frac{\s_l + \bm{\bm{\xi}}_l}{\lVert \bm{\bm{\xi}}_l \rVert_*}.
    \]
    This approximation follows directly from the triangle inequality, as the spectral norm of the noise significantly exceeds that of the signal (i.e., $\|\bm{\xi}_l\|_* \gg \|\s_l\|_*$). Consequently, the effective contribution of the signal component to this update is $\Delta \W_{\text{signal}} \asymp \frac{\s_l}{\lVert \bm{\bm{\xi}}_l \rVert_*}$. Applying the SNR condition established in Proposition~\ref{prop:snr_condition} and substituting it into the expression for the signal component:
    \[
        \lVert \Delta \W_{\text{signal}} \rVert_* \asymp \frac{\lVert \s_l \rVert_*}{c_l^{\mathrm{proxy}} \cdot \lVert \s_l \rVert_*} = \frac{1}{c_l^{\mathrm{proxy}}}= \frac{1}{c_l^{\mathrm{target}}}.
    \]
    Since $c_l^{\mathrm{target}}$ is a constant determined solely by the fixed target architecture, the effective signal strength is invariant to the proxy width. This ensures that the feature learning signal is preserved at a consistent scale relative to the target model, allowing hyperparameters tuned on small proxies to be directly transferred.
    
     {We next show necessity.
     Under DP, the noise sets the scale of the update and the signal drives feature learning, so transfer requires both to match across the two models.
     The learning rate $\eta_l$ scales the signal and the noise by the same factor, so their ratio $c_l$ does not depend on $\eta_l$.
     Therefore, if $c_l^{\mathrm{proxy}} \neq c_l^{\mathrm{target}}$, no learning rate can match the noise scale and the signal scale simultaneously, and hyperparameter transfer fails.}
\end{proof}
\begin{remark}
    For batch size transfer, we can treat the ratio $\frac{\sigma_{\text{DP}}}{B}$ as a unified term. This simplifies the derivation of scaling laws, as the effective noise level remains invariant relative to the signal.
\end{remark}

\subsection{Layer-wise Clipping Parameter}
Since the input dimension $n_0$ (e.g., embedding size) and output dimension $n_L$ (determined by the specific task) are typically fixed, the model widths at the boundaries do not scale proportionally with the hidden layers. Consequently, we cannot rely solely on noise scaling to match the SNR at the input and output layers.

To address this, given a target privacy noise level $\sigma$, we treat the layer-wise clipping coefficient as the adjustable parameter to satisfy the SNR condition in the proxy model (characterized by noise $\sigma^\prime$). Let $D_l$ and $D_l^\prime$ denote the clipping parameters for the target and proxy models, respectively. We obtain the following system of constraints, which determines the optimal layer-wise parameters:
\begin{equation}
\label{eq:scaling_squared}
\left\{
\begin{aligned}
    & \left(\frac{\sigma^\prime}{\sigma}\right)^2 \frac{D_l^\prime}{D_l}
      = \left(\frac{\sqrt{n_{l-1}} + \sqrt{ n_l}}{\sqrt{n_{l-1}^\prime} + \sqrt{n_l^\prime}}\right)^2:= f_l,\quad l\in\{1,\dots,L\}\\
    & \sum_{l=1}^{L} \frac{1}{D_l} = \sum_{l=1}^{L} \frac{1}{D_l'} = 1,
\end{aligned}
\right.
\end{equation}
The rescaled parameters $(\sigma^\prime, D_i')$ admit the closed-form solution $\sigma^\prime = \sigma / \sqrt{\sum_{i=1}^{L} \frac{1}{D_i f_i}}$, $ D_i' = D_i f_i \cdot \sum_{t=1}^{L} \frac{1}{D_t f_t}.$
Equation~\eqref{eq:scaling_squared} enforces layer-wise SNR alignment, ensuring that the proxy model effectively mimics the optimization dynamics of the target model despite the difference in width. {The LayerNorm operation in Transformers differs from standard linear layers, we present the detailed derivations in Appendix~\ref{sec:appendix_layernorm}.}

{In practice, DP-$\mu$P proceeds as follows.
We first calibrate the noise multiplier of the target model with the privacy accountant and adopt uniform clipping on the proxy.
Solving equation \eqref{eq:scaling_squared} then gives the noise multiplier of the proxy and the layer-wise clipping parameters of the target.
Finally, we grid search the learning rate on the proxy and map it to the layer-wise learning rates of the target through Lemma~\ref{lem:layerwise lr}.
Appendix~\ref{sec:derivation_example} presents this procedure step by step with a worked example.}


\section{Experiments}
\label{sec:experiments}
In this section, we evaluate the effectiveness of DP-$\mu$P in two key aspects: (i) enabling robust hyperparameter transfer across model scales, including variations in width and batch size; and (ii) achieving comparable or superior performance under differential privacy constraints. {Following the $\mu$P literature, we compare training loss across widths in the main text, and report multi-seed results on the final test accuracy in Appendix~\ref{app:utility}.}

Our experiments cover two primary settings: (i) image classification, focusing on scale transfer and gradient coordinate analysis, and (ii) text generation. We compare our approach against standard parametrization (SP) using multiple optimizers (SGD, Adam, and Muon). Unless otherwise specified, all differentially private experiments use a fixed privacy budget of $\epsilon=2$ and $\delta=10^{-5}$. {Following common practice \cite{de2022unlockinghighaccuracydifferentiallyprivate,yu2022differentiallyprivatefinetuninglanguage} , this budget covers the training of the target model and excludes the privacy cost of the learning rate search on the proxy. DP-$\mu$P therefore assumes that the proxy learning rate is either selected privately \cite{mohapatra2021roleadaptiveoptimizershonest,papernot2022hyperparametertuningrenyidifferential} or obtained from public data \cite{hod2025reallyneedpublicdata,Ponomareva_2023}; Appendix~\ref{sec:end_to_end_privacy} gives the end-to-end accounting and shows that the search adds little privacy cost.}

Furthermore, all experiments were conducted using a single NVIDIA A100 GPU. This comprehensive setup allows us to assess both the robustness and generality of DP-$\mu$P across diverse architectures, tasks, and optimization landscapes.

\subsection{Image Classification Pretraining}
We begin with experiments on CIFAR-10 to evaluate DP-$\mu$P in a controlled setting, focusing on both hyperparameter transfer and optimization performance. Our evaluation spans hyperparameter transfer across width using a 5-layer MLP and Vision Transformer (ViT), complemented by a coordinate-wise gradient analysis under different optimizers.

\begin{figure*}[!htb]
    \centering
    \includegraphics[width=0.95\linewidth]{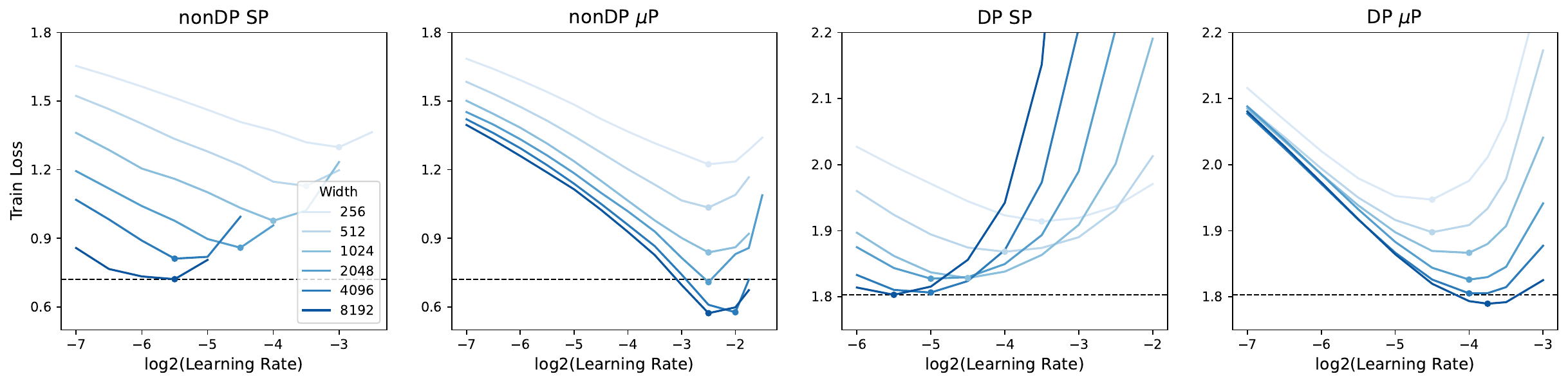}
    \caption{\textbf{Comparison between SP and $\mu$P (SGD).} We adjust the clipping coefficient and noise to satisfy the SNR condition while keeping the input dimension fixed. }
    \label{fig:mlp-sgd-loss}
\end{figure*}

 \begin{figure*}[!htb]
     \centering
     \includegraphics[width=0.95\linewidth]{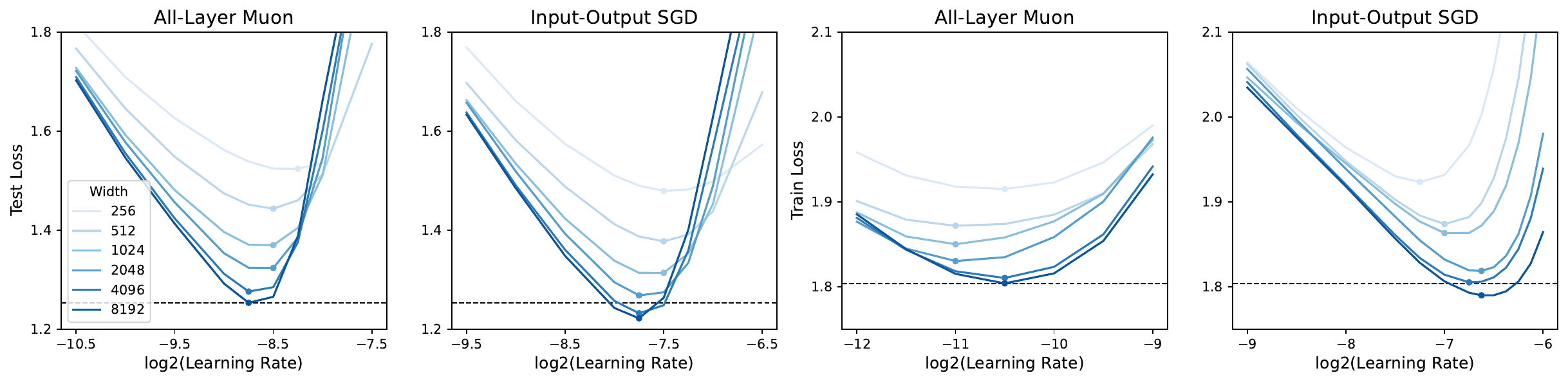}
     \caption{\textbf{Comparison between Muon and SGD-Muon.} We evaluate different update rules (Muon vs. SGD) for the Input and Output layers under the $\mu$P scaling conditions. The left two panels show the non-private setting (test loss), and the right two panels show the DP setting (training loss).}
     \label{fig:mlp-muon-loss}
 \end{figure*}

Figure~\ref{fig:mlp-sgd-loss} illustrates the training loss of a 5-layer MLP on CIFAR-10 over 10 epochs. We display the SGD and Muon results here as representative examples, and defer the results for Adam to Appendix~\ref{sec:supp_adam_muon}. Across all optimizers, $\mu$P consistently yields more stable training dynamics and lower final loss compared to standard parametrization (SP).

Regarding the input and output layers in the Muon optimizer, we deviate from the original implementation, which typically employs Adam updates. Instead, we apply either Muon or SGD updates. This modification is necessary because the standard Adam update fails to exhibit hyperparameter transfer in this context, even in non-private settings. {Figure~\ref{fig:mlp-muon-loss} compares these two choices in both the non-private and the DP setting. For the non-private results, we report test loss instead of training loss, because Muon overfits significantly in the non-private regime.
}

\begin{figure*}[!htb]
    \centering
    \includegraphics[width=0.7125\linewidth]{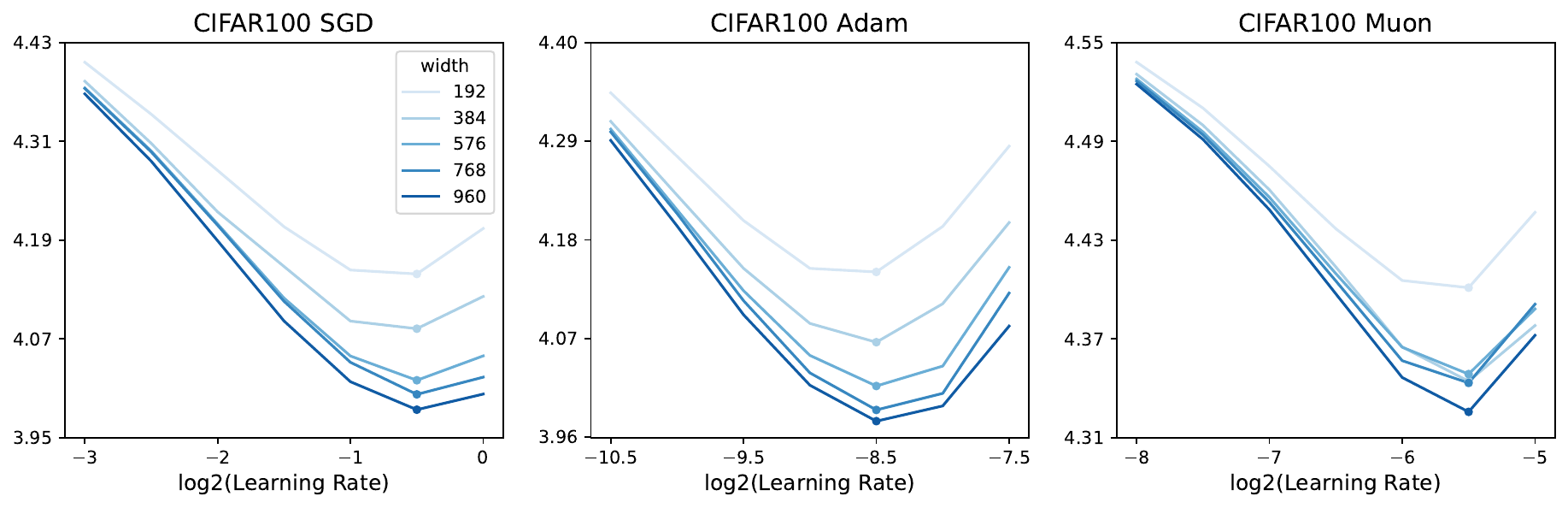}
    \caption{\textbf{DP-$\mu$P ViT on CIFAR-100.} For Muon, the input and output layers are updated using SGD.}
    \label{fig:vit-cifar100}
\end{figure*}
Figure \ref{fig:vit-cifar100} displays the pretraining loss curves for ViT models of increasing width on CIFAR-100, with corresponding results for CIFAR-10 deferred to Appendix~\ref{app:vit-sp}. Across all widths, $\mu$P enables effective hyperparameter transfer, allowing larger models to be trained using the same learning rate as smaller ones in the DP setting. Furthermore, larger ViTs consistently achieve lower training loss, provided the SNR condition is satisfied.

We also extend our evaluation to batch size transferability. In this setting, we fix the model width and scale the noise multiplier $\sigma$ proportionally to the batch size. Crucially, to ensure a rigorous comparison of optimization trajectories, we fix the total number of update steps (iterations) across all runs, rather than the number of epochs. The detailed results confirming robust transfer under these conditions are provided in \cref{app:batch-size}.

\subsection{Coordinate Check}
\begin{figure*}[!htb]
    \centering
    \includegraphics[width=1\linewidth]{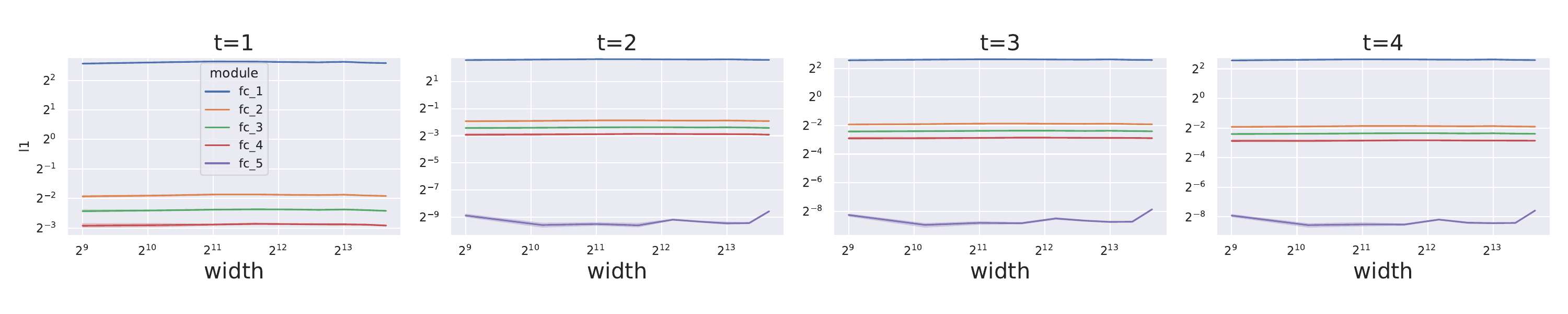}
    \vspace{-0.2in}
    \caption{\textbf{Coordinate check for MLP (DP-$\mu$P).} The evolution of activation $l_1$ norms across different widths shows consistent scaling, confirming correct implementation.}
    \label{fig:mlp-coord}
\end{figure*}

\begin{figure*}[!htb]
    \centering
    \includegraphics[width=1.0\linewidth]{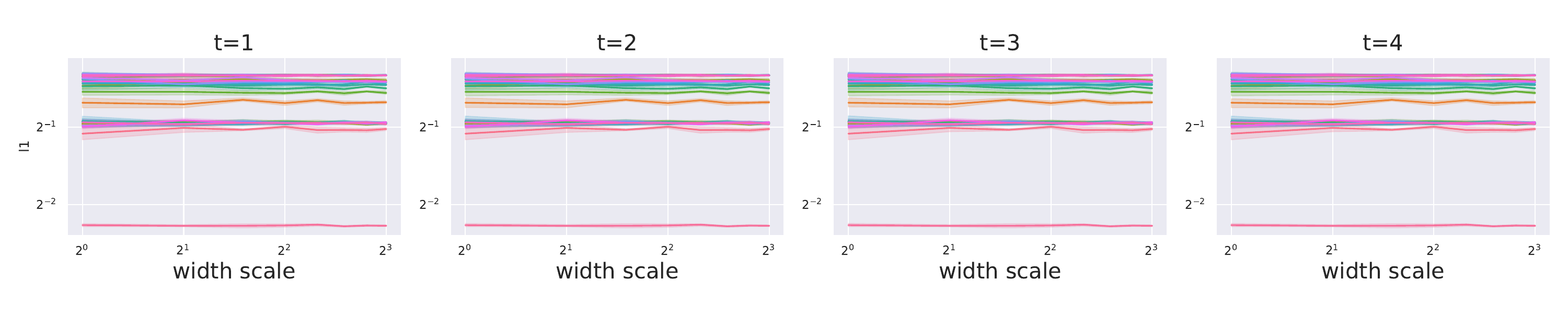}
    \vspace{-0.2in}
    \caption{\textbf{Coordinate check for ViT (DP-$\mu$P).} Similar to the MLP case, the Vision Transformer maintains stable activation scales across varying widths.}
    \label{fig:vit-coord}
\end{figure*}

While the theoretical foundations of DP-$\mu$P are well-established, practical implementation remains error-prone. To support correctness, we adopt the coordinate checking procedure introduced by~\cite{yang2022tensor}. As illustrated in Figures~\ref{fig:mlp-coord} and \ref{fig:vit-coord}, we track the $l_1$ norm of feature activations for specific coordinates over the initial training steps, sweeping model width across a broad range. A correct $\mu$P implementation preserves the scale of these activations across varying widths. For comparison, coordinate check results for standard parametrization (SP) are provided in Appendix~\ref{app:coord-check-sp}, where activation norms typically explode as width increases.

\subsection{Text Generation Pretraining}
\label{sec:exp gpt}
\begin{figure*}[!htb]
    \centering
    \includegraphics[width=0.95\linewidth]{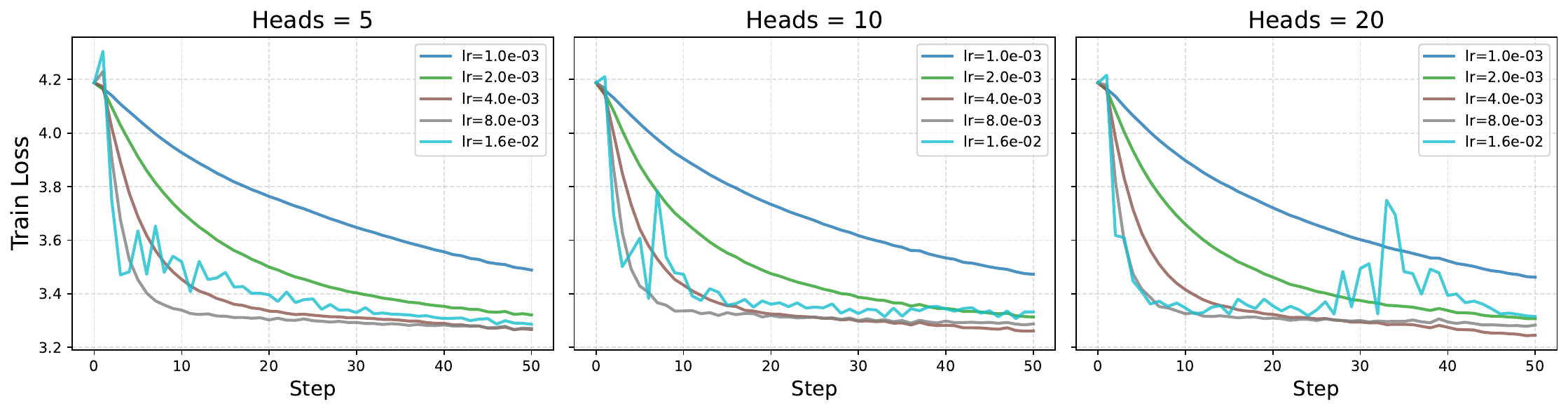}
    \caption{\textbf{GPT-2 Pretraining with DP-$\mu$P on Shakespeare Character-Level Task.} Loss curves across varying widths (controlled by head counts) exhibit consistent ranking and dynamics. This consistency indicates that the optimal learning rate remains invariant across model sizes, demonstrating successful hyperparameter transfer.}
    \label{fig:gpt2-loss}
\end{figure*}

To validate our approach on generative tasks, we conduct pretraining experiments using the GPT-2 architecture on the Shakespeare Character dataset. Our target configuration is a GPT-2 Large model (36 layers, 20 heads, $d_{model}=1280$) with 787.8M parameters. To maximize search efficiency, we determine the optimal learning rate on a small proxy model (5 heads, $d_{model}=320$, 76.0M parameters), which yields a 90.35\% reduction in theoretical computational cost. { Empirically, on a single NVIDIA A100 GPU, this reduces the wall-clock execution time per training run from approximately 30 minutes to 11 minutes.} Additionally, we evaluate an intermediate model (10 heads, 227.3M parameters) to corroborate the consistency of the loss landscape across widths. As shown in Figure~\ref{fig:gpt2-loss}, the training loss curves exhibit remarkably consistent patterns across all three model scales for each learning rate, demonstrating successful zero-shot hyperparameter transfer.

\section{Discussion}
{Maximal update parametrization ($\mu$P) has inspired numerous extensions across various architectures and training regimes \cite{blake2025umupunitscaledmaximalupdate,yang2023tensorprogramsvifeature,zheng2025scalingdiffusiontransformersefficiently}.} Our work introduces DP-$\mu$P, a theoretically grounded and empirically validated framework for hyperparameter transfer in differentially private deep learning. By bridging the gap between $\mu$P and differential privacy, we resolve the tension between gradient clipping and feature scaling. Specifically, our derivation of layer-wise learning rate scaling rules ensures that the Signal-to-Noise Ratio (SNR) remains consistent across model scales, a critical requirement often overlooked in standard DP optimization.

Empirically, we validate this theoretical framework across diverse settings, including MLP, ViT, and GPT-2 architectures. Our results demonstrate that optimal hyperparameters (e.g., learning rate and clipping threshold) identified on small, inexpensive proxy models transfer zero-shot to large-scale targets. {This capability provides a critical alternative to established DP hyperparameter tuning strategies \cite{de2022unlockinghighaccuracydifferentiallyprivate, papernot2022hyperparametertuningrenyidifferential}, which typically require repeated search iterations or rely heavily on auxiliary public datasets, thereby drastically reducing the computational costs.}

\section{Limitations and Future Work}
\label{sec:limitations}
While our work establishes robust scaling laws for width and batch size across MLPs, ViTs, and GPT-style transformers, we do not currently address depth scaling. Extending our framework to deeper networks presents a fundamental challenge, as the accumulation of privacy noise across successive layers naturally tends to violate the layer-wise SNR condition required for stable learning. Future work is needed to either derive a depth-scaling mechanism that preserves this SNR or to rigorously establish a theoretical impossibility result.



Finally, our current analysis operates under a noise-dominant regime, an assumption that holds in typical differentially private training where gradient clipping and noise injection often overwhelm the signal.
{With extremely large batch sizes, such as those used in large-scale private pretraining \cite{sinha2025vaultgemmadifferentiallyprivategemma}, the noise-to-signal ratio becomes small and the scaling reduces to the clipping-only rules in \Cref{tab:lr and init muP}. A valuable direction for future research would be to extend this analysis to the transition regime, where signal and noise magnitudes are comparable, to understand if the scaling laws require adjustment when the privacy constraint is relaxed.}

\newpage

\bibliography{example_paper}
\bibliographystyle{plain}

\appendix

\section{Mathematical Derivations} 
\label{section: proof}

\subsection{Handling Bias and Nonlinearities}
\label{sec:bias and nonlinearities}
To justify the simplification in the main text, we analyze the magnitude of the bias terms and the effect of nonlinear activation functions. We show that they do not alter the order of magnitude of the hidden states or gradients with respect to the width $n$.

\subsubsection{Bias Terms.}
Let $\b_l \in \mathbb{R}^{n_l}$ be the bias vector at layer $l$. The pre-activation is given by $\a_l = \a_{l-1} \W_l  + \b_l$. following the perspective in modern parametrization literature \citep{yang2023spectral}, we view the bias vector as a weight matrix $\b_l \in \mathbb{R}^{n_l \times 1}$ connecting an auxiliary layer with width $1$ and a constant output of $1$ to the $l$-th hidden layer.

This unification allows us to apply our spectral scaling analysis directly to the bias $\b_l$. A key property utilized here is that for any vector $v$, its spectral norm is equivalent to its Euclidean norm:
\begin{equation}
    \|\b_l\|_* = \|\b_l\|_2
\end{equation}
Under the spectral scaling condition, treating $\b_l$ as a weight matrix prescribes $\|\b_l\|_2 = \Theta(\sqrt{n_l})$. Furthermore, this formulation naturally accommodates the SNR condition under layer-wise clipping, where the weight matrix and bias share the same clipping coefficient and noise scale. Assuming proportional width scaling where the aspect ratio between layers is preserved (i.e., $n_{l-1}/n_{l-1}^\prime = n_l/n_l^\prime$), the scaling factor simplifies as follows:
\begin{equation}
    \frac{\sigma^\prime}{\sigma} \sqrt{\frac{D_l^\prime}{D_l}}
      = \frac{\sqrt{n_{l-1}} + \sqrt{ n_l}}{\sqrt{n_{l-1}^\prime} + \sqrt{n_l^\prime}} 
      = \frac{\sqrt{n_l}}{\sqrt{n_l^\prime}}
\end{equation}
The last equality holds because the term $(\sqrt{n_{l-1}/n_l} + 1)$ factors out from both the numerator and the denominator.

\subsubsection{Nonlinear Activations}
Following the analysis in prior literature \cite{yang2023spectral}, we adopt the assumption that the activation function $\phi(\cdot)$ preserves the order of the signal magnitude:
\begin{equation}
    \|\phi(\a_l)\|_2 = \Theta(\|\a_l\|_2)
\end{equation}
This assumption is justified by the design of standard activation functions (e.g., ReLU, GELU), which are constructed to take order-one inputs and yield outputs that neither explode nor uniformly vanish. Consequently, the introduction of nonlinearity does not alter the asymptotic scaling of the network width, allowing our spectral analysis to proceed effectively under the linear approximation.

\subsection{Proof of Lemma~\ref{lem:layerwise lr}}
\begin{lemma}
\label{app_lem:layerwise lr}
For any optimizer gradient update $\g_l$, satisfying the spectral scaling condition for feature learning is equivalent to setting the layer-wise learning rate $\eta_l$ as:
\begin{equation}
    \eta_l = \Theta\left(\frac{\sqrt{n_l/n_{l-1}}}{\|\g_l\|_*}\right).
\end{equation}
\end{lemma}
In this section, we provide a simplified proof of Lemma~\ref{app_lem:layerwise lr} (referenced as Lemma~\ref{lem:layerwise lr} in the main text) to illustrate the core mechanism. For the complete rigorous derivation and additional theoretical details, we refer readers to \cite{yang2023spectral}.

\begin{definition}[Desideratum 1 in \cite{yang2023spectral}]
\label{def:feature_learning} 
Let $\a_l(x) \in \mathbb{R}^{n_l}$ denote the features of input $x$ at layer $l$ of a neural network, and let $\Delta \a_l(x) \in \mathbb{R}^{n_l}$ denote their change after a gradient step. We desire that:
\begin{equation}
    \|\a_l\|_2 = \Theta(\sqrt{n_l}) \quad \text{and} \quad \|\Delta \a_l\|_2 = \Theta(\sqrt{n_l}),
\end{equation}
for layers $l = 1, \dots, L-1$.
\end{definition}

\begin{proof}
Recall that the weight update at step $t$ is given by $\Delta \W_l = -\eta_l \cdot \g_l$. Taking the spectral norm of both sides, we have:
\begin{equation}
    \|\Delta \W_l\|_* = \eta_l \|\g_l\|_*.
\end{equation}
To ensure feature learning in the limit of infinite width, the spectral scaling condition requires the weight update to scale as:
\begin{equation}
    \|\Delta \W_l\|_* = \Theta\left(\sqrt{\frac{n_l}{n_{l-1}}}\right).
\end{equation}
This requirement stems from the property that the change in output activations satisfies $\|\Delta \a_l\|_2 \asymp \|\Delta \W_l\|_* \cdot \|\a_{l-1}\|_2$ when the input $\a_{l-1}$ aligns with the top singular vector of the weight update $\Delta \W_l$. Since $\|\a_{l-1}\|_2 = \Theta(\sqrt{n_{l-1}})$, ensuring $\|\Delta \W_l\|_* = \Theta(\sqrt{n_l/n_{l-1}})$ guarantees that the feature updates scale comparably to the activations themselves ($\|\Delta \a_l\|_2 = \Theta(\sqrt{n_l})$), enabling stable feature learning.

Substituting this condition into the update equation, we solve for $\eta_l$:
\begin{equation}
    \eta_l \|\g_l\|_* = \Theta\left(\sqrt{\frac{n_l}{n_{l-1}}}\right) \implies \eta_l = \Theta\left(\frac{\sqrt{n_l/n_{l-1}}}{\|\g_l\|_*}\right).
\end{equation}
This completes the proof.
\end{proof}

\subsection{Derivation of Layer-wise Learning Rate Scaling Rules}
\label{app:lr_derivation}
\begin{table*}[h]
  \centering
  \caption{Summary of layer-wise learning rate scaling rules for $\mu$P under different privacy settings. This table restates \Cref{tab:lr and init muP} so that the derivation in this section can be read on its own.}
  \renewcommand{\arraystretch}{2.0}
  \setlength{\tabcolsep}{8pt}
  \begin{tabular}{lccc}
    \toprule
    \textbf{Optimizer} & \textbf{Non-Private} & \textbf{Clipping Only} & \textbf{Differentially Private (DP)} \\
    \midrule
    SGD  & $\dfrac{n_{l}}{n_{l-1}}$ 
         & $\sqrt{\dfrac{n_l}{n_{l-1}}}$ 
         & $\dfrac{\sqrt{n_l / n_{l-1}}}{\frac{\sigma_{\text{DP}}}{B}(\sqrt{n_l}+\sqrt{n_{l-1}})}$ \\
    
    Adam & $\dfrac{1}{n_{l-1}}$     
         & $\dfrac{1}{n_{l-1}}$          
         & $\dfrac{\sqrt{n_l / n_{l-1}}}{\sqrt{n_l}+\sqrt{n_{l-1}}}$ \\
    
    Muon & $\sqrt{\dfrac{n_l}{n_{l-1}}}$ 
         & $\sqrt{\dfrac{n_l}{n_{l-1}}}$ 
         & $\sqrt{\dfrac{n_l}{n_{l-1}}}$ \\ 
    \bottomrule
  \end{tabular}
  \label{app tab:lr_summary_app}
\end{table*}

We derive the optimal layer-wise learning rates $\eta_l$ by ensuring that the spectral norm of the weight update satisfies the Feature Learning condition:
\begin{equation}
    \|\Delta\W_l\|_* = \Theta\left(\sqrt{\frac{n_l}{n_{l-1}}}\right).
\end{equation}
First, we recall the assumptions and algebraic facts used in our derivations.

\begin{fact}
\label{app_fact:rank1_norm}
For rank-1 matrices (such as per-sample gradients in standard networks), the spectral norm is equal to the Frobenius norm: $\|\nabla_{\W_l}\mathcal{L}_i\|_* = \|\nabla_{\W_l}\mathcal{L}_i\|_F$.
\end{fact}

\begin{assumption}
\label{app_assump:aligned}
We assume that the per-sample gradients are sufficiently aligned across the batch, such that the spectral norm of the sum scales with the sum of the spectral norms:
\begin{equation}
    \left\|\frac{1}{B}\sum_{i=1}^B g(\p_i)\right\|_* = \Theta\left(\frac{1}{B}\sum_{i=1}^B\|g(\p_i)\|_*\right).
\end{equation}
\end{assumption}

\begin{fact}
\label{app_fact:sign_rank1}
The sign operator preserves the rank-1 structure for outer products: $\text{sign}(uv^\top)=\text{sign}(u)\text{sign}(v)^\top$. Consequently, $\|\text{sign}(uv^\top)\|_* = \|\text{sign}(uv^\top)\|_F = \Theta(\sqrt{n_{l} n_{l-1}})$.
\end{fact}

\subsubsection{Non-Private Case (Standard $\mu$P)}
In the standard non-private setting, we derive the scaling of the gradient spectral norms for SGD and Adam.

\textbf{SGD:}
Using the definition of the update and applying our alignment assumption:
\begin{equation}
\begin{aligned}
    \|g(\p_l)\|_* &= \left\|\frac{1}{B}\sum_{i=1}^B {\nabla_{\W_l} \mathcal{L}_i}\right\|_* \overset{\text{(A.2)}}{=} \Theta\left( \frac{1}{B}\sum_{i=1}^B {\|\nabla_{\W_l} \mathcal{L}_i\|_*}\right) \\
    &\overset{\text{(A.1)}}{=} \Theta\left( \frac{1}{B}\sum_{i=1}^B \sqrt{\frac{n_{l-1}}{n_{l}}}\right) = \Theta\left(\sqrt{\frac{n_{l-1}}{n_{l}}} \right).
\end{aligned}
\end{equation}

\textbf{SignSGD / Adam:}
For sign-based optimizers, the update magnitude depends on the full dimension of the matrix due to the element-wise operation:
\begin{equation}
\begin{aligned}
    \|g(\p_l)\|_* &= \left\|\frac{1}{B}\sum_{i=1}^B \text{sign}({\nabla_{\W_l} \mathcal{L}_i})\right\|_* \overset{\text{(A.2)}}{=} \Theta\left( \frac{1}{B}\sum_{i=1}^B {\|\text{sign}(\nabla_{\W_l} \mathcal{L}_i)\|_*}\right) \\
    &\overset{\text{(A.3)}}{=} \Theta\left( \frac{1}{B}\sum_{i=1}^B {\|\text{sign}(\nabla_{\W_l} \mathcal{L}_i)\|_F}\right) = \Theta\left( \frac{1}{B}\sum_{i=1}^B \sqrt{{n_{l-1}}{n_{l}}}\right) = \Theta\left(\sqrt{{n_{l-1}}{n_{l}}} \right).
\end{aligned}
\end{equation}

\textbf{Muon:}
The Muon optimizer updates weights using an orthogonalized momentum. The update rule is defined as:
\begin{equation}
    \Delta \W_l = -\eta_l \cdot g(\p_l) := -\eta_l \cdot \mathcal{O}(m(\p_l)),
\end{equation}
where $m(\p_l)$ is the momentum accumulator for the gradient, and $\mathcal{O}(\cdot)$ denotes the orthogonalization operator. This operator projects the momentum matrix onto the manifold of orthogonal matrices by solving the following optimization problem:
\begin{equation}
    \mathcal{O}(\bm{M}) = \operatorname*{argmin}_{\bm{U}: \bm{U}^\top \bm{U} = I} \| \bm{M} - \bm{U} \|_F.
\end{equation}
The constraint $\bm{U}^\top \bm{U} = I$ explicitly controls the spectral properties of the update. By definition, any matrix $\bm{U}$ satisfying this constraint has all singular values equal to 1. Consequently, the spectral norm of the update direction is invariant to the choice of momentum or the presence of privacy noise:
\begin{equation}
    \|g(\p_l)\|_* \equiv \|\mathcal{O}(m(\p_l))\|_* = 1 = \Theta(1).
\end{equation}
To satisfy the feature learning condition $\|\Delta \W_l\|_* = \Theta(\sqrt{n_l/n_{l-1}})$, we apply the relationship $\eta_l = \|\Delta \W_l\|_* / \|g(\p_l)\|_*$ using the spectral norms derived above. These scaling rules correspond directly to the first column of Table~\ref{app tab:lr_summary_app}.

\subsubsection{Clipping-Only Case}
In this regime, we assume per-sample gradient clipping is active, but noise injection is negligible. The pre-conditioned gradient is defined as $g(p_l)$.

Note that for SignSGD/Adam and Muon, the clipping operation (which scales the gradient by a scalar) does not fundamentally alter the rank or the spectral properties of the update matrix; thus, their scaling orders remain unchanged compared to the non-private setting. However, for SGD, the normalization by the Frobenius norm significantly alters the scaling behavior, as derived below.

\textbf{SGD:}
The update is defined as the average of clipped per-sample gradients. By invoking the sufficient alignment assumption (Assumption~\ref{app_assump:aligned}), we derive:
\begin{align}
\|g(\p_l)\|_* &= \left\|\frac{1}{B}\sum_{i=1}^B \frac{\nabla_{\W_l}\mathcal{L}_i}{\sqrt{D}\|\nabla_{\W_l}\mathcal{L}_i\|_F}\right\|_* \nonumber \overset{\text{(A.2)}}{=} \Theta\left( \frac{1}{B}\sum_{i=1}^B {\frac{\|\nabla_{\W_l} \mathcal{L}_i\|_*}{{\sqrt{D}\|\nabla_{\W_l}\mathcal{L}_i\|_F}}}\right) \\ & \nonumber \overset{\text{(A.1)}}{=} \Theta\left( \frac{1}{B}\sum_{i=1}^B \frac{1}{\sqrt{D}} \right) = \Theta(1),
\end{align}
where $D$ denotes the number of layers, which is independent of the model width.

\subsubsection{Differentially Private Case (Noise-Dominated)}
In the DP setting, the update is dominated by the injected Gaussian noise $\bm{\bm{\xi}} \sim \mathcal{N}(0, I_{n_l \times n_{l-1}})$. We utilize the result from Random Matrix Theory that the spectral norm of a random Gaussian matrix of size $n_l \times n_{l-1}$ scales as $\sqrt{n_l} + \sqrt{n_{l-1}}$.

\textbf{DP-SGD:}
The gradient is dominated by noise scaled by $\frac{\sigma_{\text{DP}}}{B}$:
\begin{equation}
    \|g(\p_l)\|_* \asymp \left\| \frac{\sigma_{\text{DP}}}{B} \bm{\bm{\xi}} \right\|_* = \Theta\left( \frac{\sigma_{\text{DP}}}{B} (\sqrt{n_l} + \sqrt{n_{l-1}}) \right).
\end{equation}

\textbf{DP-Adam:}
In the noise-dominated regime, the update geometry of Adam is governed by the injected noise, and the spectral norm of the update direction scales as:
\begin{equation}
    \|g(\p_l)\|_* = \Theta(\sqrt{n_l} + \sqrt{n_{l-1}}).
\end{equation}
This differs from the clipping-only case, where the norm scales as $\sqrt{n_l n_{l-1}}$, because the noise does not align across coordinates as the signal gradients do.

\textbf{DP-Muon:}
As in the non-private case, the orthogonalization sets all singular values of the momentum to one, regardless of the privacy noise:
\begin{equation}
    \|g(\p_l)\|_* = \Theta(1).
\end{equation}
In practice, Muon approximates $\mathcal{O}(\cdot)$ with a few Newton--Schulz iterations, which act only on the singular values and bring them close to one, so the width dependence is unchanged.

\subsection{Boundary Condition at the Input Layer}
\label{sec:input_layer_scaling}
Unlike hidden layers, where the input signal strength depends on the width of the preceding layer $n_{l-1}$, the input layer receives fixed data samples $\x \in \mathbb{R}^{n_0}$. The feature norm at the first layer is bounded by:
\begin{equation}
    \|\a_1\| \le \|\W_{in}\|_* \cdot \|\x\|.
\end{equation}
Standard data preprocessing in deep learning (e.g., image whitening or standardization) typically ensures that input features have zero mean and unit variance per dimension. Consequently, the norm of the input vector scales with the square root of the input dimension:
\begin{equation}
    \mathbb{E}[\|\x\|^2] = \sum_{j=1}^{n_0} \mathbb{E}[x_j^2] = n_0 \cdot \Theta(1) \implies \|\x\| = \Theta(\sqrt{n_0}).
\end{equation}
Given that $n_0$ is fixed for a specific task, we do not explicitly distinguish this scaling factor in our main notation. However, we note that if the data is standardized differently (e.g., non-unit variance), the learning rate scale for the input layer would need to be adjusted accordingly.

\subsection{SNR for Layer Normalization Parameters}
For the affine parameters in Layer Normalization (specifically the gain vector $\gamma \in \mathbb{R}^{n_l}$), the operation corresponds to element-wise multiplication. This is mathematically equivalent to matrix multiplication by a diagonal matrix $\W_{\mathcal{L}_{\text{norm}}} = \text{diag}(\gamma)$, where $\W_{\mathcal{L}_{\text{norm}}} \in \mathbb{R}^{n_l \times n_l}$.

Consequently, the spectral norm of the LayerNorm gradient update matrix, denoted as $g(p_{\mathcal{L}_{\text{norm}}})$, behaves differently from dense weight matrices. The spectral norm of a diagonal matrix is simply the maximum absolute value of its diagonal entries (the infinity norm of the vector):
\begin{equation}
    \|g(\p_{\mathcal{L}_{\text{norm}}})\|_* = \|\text{diag}(g(\gamma))\|_* = \max_{i \in \{1, \dots, n_l\}} |g(\gamma)_i|.
\end{equation}
Unlike dense matrices where the spectral norm typically scales with $\sqrt{n_l}$, the scaling here is governed by \textit{Extreme Value Theory}. Assuming the gradient entries (or the injected DP noise) are independent and identically distributed with sub-Gaussian tails, the expected maximum of $n_l$ variables scales logarithmically:
\begin{equation}
    \|g(\p_{\mathcal{L}_{\text{norm}}})\|_* = \Theta\left(\sqrt{\ln n_l}\right).
\end{equation}
To maintain the SNR condition, we require the effective noise magnitude to scale consistently. This implies the following relationship:
\begin{equation}
    \frac{\sigma \sqrt{D_k}}{B}\sqrt{\ln (n_k)} =\frac{\sigma^\prime \sqrt{D_k^\prime}}{B}\sqrt{\ln (n_k^\prime)} 
    \iff \left(\frac{\sigma^\prime}{\sigma}\right)^2 \frac{D_k^\prime}{D_k} = \frac{\ln(n_k)}{\ln(n_k^\prime)}, \quad \forall k \in \mathcal{L}_{\text{norm}}.
\end{equation}

\subsection{Scaling Rules for Transformers with LayerNorm}
\label{sec:appendix_layernorm}

In the main text, Equation~\eqref{eq:scaling_squared} establishes the layer-wise constraints for standard linear networks. Here, we formally extend this system to Transformer architectures by explicitly accounting for normalization layers. Let $\mathcal{L}_{\text{linear}}$ denote the set of indices for standard linear layers, and $\mathcal{L}_{\text{norm}}$ denote the set for normalization layers.

\begin{remark}
Note that the operation of LayerNorm in Transformers is analogous to diagonal matrix multiplication. However, driven by extreme value theory, the spectral norm of its gradient scales with $\sqrt{\ln n}$ rather than the $\sqrt{n}$ scaling observed in linear layers. Specifically, the expanded system of constraints is given by:
\begin{equation}
\label{eq:scaling_system_exact_snr}
\left\{
\begin{aligned}
    & \left(\frac{\sigma^\prime}{\sigma}\right)^2 \frac{D_l^\prime}{D_l}
      = \left( \frac{\sqrt{n_{in}^{(l)}} + \sqrt{n_{out}^{(l)}}}{\sqrt{n_{in}^{(l)\prime}} + \sqrt{n_{out}^{(l)\prime}}} \right)^2,
      && \forall l \in \mathcal{L}_{\text{linear}}, \\[8pt]
    & \left(\frac{\sigma^\prime}{\sigma}\right)^2 \frac{D_k^\prime}{D_k}
      = \frac{\ln(n_k)}{\ln(n_k^\prime)},
      && \forall k \in \mathcal{L}_{\text{norm}}, \\[8pt]
    & \sum_{j \in \mathcal{L}_{\text{linear}} \cup \mathcal{L}_{\text{norm}}} \frac{1}{D_j}
      = \sum_{j \in \mathcal{L}_{\text{linear}} \cup \mathcal{L}_{\text{norm}}} \frac{1}{D_j^\prime} = 1.
\end{aligned}
\right.
\end{equation}
\end{remark}

\section{Empirical Support for Assumption \ref{assum:alignment}}
\label{sec:empirical_alignment}
{In this section, we provide empirical validation for Assumption \ref{assum:alignment}:
$$\left\|\frac{1}{B}\sum_{i=1}^B g(\p_i)\right\|_* = \Theta\left(\frac{1}{B}\sum_{i=1}^B\|g(\p_i)\|_*\right)$$
We evaluate a 5-layer Multi-Layer Perceptron (MLP) trained with DP-SGD and DP-Adam. DP-Muon is omitted from this analysis, as its preconditioning operation inherently maps updates to matrices with a unit spectral norm. Throughout the experiments, we maintain a fixed noise multiplier $\sigma$ and learning rate, while systematically varying the batch size $B \in \{125, 250, 500, 1000\}$ and the network width $\in \{256, 512, 1024\}$.

\begin{table}[H]
  \caption{Empirical spectral norm of the mini-batch average gradients for DP-SGD across varying batch sizes and network widths.}
  \label{tab:sgd-spectral}
  \centering
  \setlength{\tabcolsep}{4pt}
  \renewcommand{\arraystretch}{1.15}
  \resizebox{\textwidth}{!}{%
  \begin{tabular}{r |cccc|cccc|cccc}
    \toprule
     & \multicolumn{4}{c}{width = 256} & \multicolumn{4}{c}{width = 512} & \multicolumn{4}{c}{width = 1024} \\
    \cmidrule(lr){2-5} \cmidrule(lr){6-9} \cmidrule(lr){10-13}
     $B=$ & 125 & 250 & 500 & 1000 & 125 & 250 & 500 & 1000 & 125 & 250 & 500 & 1000 \\
    \midrule
    Input         & 1.13 & 0.56 & 0.28 & 0.14 & 1.25 & 0.62 & 0.31 & 0.16 & 1.40 & 0.70 & 0.35 & 0.17 \\
    First Hidden  & 0.51 & 0.25 & 0.13 & 0.06 & 0.72 & 0.36 & 0.18 & 0.09 & 1.03 & 0.51 & 0.26 & 0.13 \\
    Second Hidden & 0.50 & 0.25 & 0.13 & 0.06 & 0.72 & 0.36 & 0.18 & 0.09 & 1.01 & 0.51 & 0.26 & 0.13 \\
    Third Hidden  & 0.51 & 0.25 & 0.13 & 0.07 & 0.72 & 0.36 & 0.18 & 0.09 & 1.02 & 0.51 & 0.25 & 0.13 \\
    Output        & 0.30 & 0.16 & 0.08 & 0.04 & 0.39 & 0.20 & 0.10 & 0.06 & 0.56 & 0.28 & 0.14 & 0.07 \\
    \bottomrule
  \end{tabular}%
  }
\end{table}

\begin{table}[H]
  \caption{Empirical spectral norm of the mini-batch average gradients for DP-Adam across varying batch sizes and network widths.}
  \label{tab:adam-spectral}
  \centering
  \setlength{\tabcolsep}{4pt}
  \renewcommand{\arraystretch}{1.15}
  \resizebox{\textwidth}{!}{%
  \begin{tabular}{r|cccc|cccc|cccc}
    \toprule
     & \multicolumn{4}{c}{width = 256} & \multicolumn{4}{c}{width = 512} & \multicolumn{4}{c}{width = 1024} \\
    \cmidrule(lr){2-5} \cmidrule(lr){6-9} \cmidrule(lr){10-13}
     $B=$ & 125 & 250 & 500 & 1000 & 125 & 250 & 500 & 1000 & 125 & 250 & 500 & 1000 \\
    \midrule
    Input         & 71.05 & 71.37 & 71.22 & 70.86 & 77.51 & 77.70 & 77.85 & 77.50 & 87.65 & 87.47 & 87.14 & 86.90 \\
    First Hidden  & 31.52 & 31.46 & 31.57 & 31.52 & 44.86 & 44.57 & 45.15 & 44.94 & 63.45 & 63.48 & 63.44 & 63.71 \\
    Second Hidden & 30.77 & 31.80 & 32.05 & 31.43 & 45.21 & 44.79 & 44.64 & 44.81 & 64.07 & 63.81 & 63.86 & 63.59 \\
    Third Hidden  & 31.96 & 31.82 & 31.41 & 31.55 & 44.66 & 44.72 & 44.82 & 44.93 & 63.52 & 63.70 & 63.55 & 63.93 \\
    Output        & 18.64 & 18.52 & 18.09 & 19.00 & 24.62 & 24.60 & 24.79 & 25.72 & 34.20 & 35.04 & 34.22 & 34.20 \\
    \bottomrule
  \end{tabular}%
  }
\end{table}

As established in our theoretical analysis, the spectral norm of the DP-SGD gradient scales as:
\begin{equation}
    \|g(\p_l)\|_* \propto \frac{\sigma}{B}\left(\sqrt{n_l} + \sqrt{n_{l-1}}\right)
\end{equation}
where $\sigma$ is the noise multiplier, $B$ is the batch size, and $n_l, n_{l-1}$ denote the input and output dimensions of layer $l$, respectively. Table~\ref{tab:sgd-spectral} empirically confirms this prediction: for a given layer and width (along each row), the spectral norm approximately halves as the batch size $B$ doubles, verifying the expected $1/B$ dependence. Furthermore, across the three width configurations, the values grow proportionally to $\sqrt{n_l} + \sqrt{n_{l-1}}$. Together, these observations strongly validate the predicted scaling behavior for DP-SGD. For DP-Adam, our analysis indicates that the spectral norm scales as:
\begin{equation}
    \|g(\p_l)\|_* \propto \left (\sqrt{n_l} + \sqrt{n_{l-1}}\right)
\end{equation}
Table~\ref{tab:adam-spectral} corroborates this behavior. The empirical spectral norm remains remarkably stable across different batch sizes for any given layer and width combination, demonstrating its independence from $B$. Simultaneously, it correctly scales proportionally to $\sqrt{n_l} + \sqrt{n_{l-1}}$ as the network width increases.}

{We further repeat the check for DP-Adam on a GPT-2 model trained on Shakespeare-char, varying the width in $\{320, 640, 1280\}$ and the batch size $B \in \{64, 128, 256\}$.
As shown in Table~\ref{tab:gpt-adam-spectral}, the spectral norms are again independent of the batch size and grow with $\sqrt{n_l} + \sqrt{n_{l-1}}$, confirming that the assumption also holds for Transformers.
Moreover, DP-$\mu$P does not require the alignment to hold exactly: the assumption enters our analysis only through the signal norm, so it suffices that the degree of alignment is comparable between the proxy and the target model.}

\begin{table}[H]
  \caption{Empirical spectral norm of the mini-batch average gradients for DP-Adam on GPT-2 across varying batch sizes and network widths.}
  \label{tab:gpt-adam-spectral}
  \centering
  \small
  \setlength{\tabcolsep}{4pt}
  \renewcommand{\arraystretch}{1.15}
  \begin{tabular}{r|ccc|ccc|ccc}
    \toprule
     & \multicolumn{3}{c}{width = 320} & \multicolumn{3}{c}{width = 640} & \multicolumn{3}{c}{width = 1280} \\
    \cmidrule(lr){2-4} \cmidrule(lr){5-7} \cmidrule(lr){8-10}
     $B=$ & 64 & 128 & 256 & 64 & 128 & 256 & 64 & 128 & 256 \\
    \midrule
    Block 5 QKV  & 48.43 & 48.38 & 48.40 & 68.58 & 68.70 & 68.96 & 97.29 & 97.69 & 97.67 \\
    Block 5 FC1  & 53.06 & 52.87 & 53.71 & 75.45 & 75.86 & 76.01 & 107.15 & 106.98 & 107.16 \\
    Block 5 FC2  & 53.05 & 53.73 & 53.10 & 75.65 & 75.36 & 75.57 & 107.05 & 107.52 & 107.52 \\
    Block 10 QKV & 48.39 & 48.36 & 47.99 & 69.27 & 68.85 & 68.88 & 97.46 & 97.68 & 97.71 \\
    Block 10 FC1 & 53.17 & 52.91 & 53.08 & 75.44 & 75.97 & 75.36 & 107.36 & 106.92 & 106.91 \\
    Block 10 FC2 & 53.27 & 53.20 & 53.36 & 75.60 & 75.44 & 75.58 & 107.07 & 106.94 & 106.94 \\
    Head         & 26.05 & 25.50 & 25.32 & 32.75 & 33.01 & 33.23 & 44.09 & 43.04 & 42.98 \\
    \bottomrule
  \end{tabular}
\end{table}

\section{Supplementary Experiments}

\subsection{Additional MLP Results for Adam}
 \label{sec:supp_adam_muon}
\begin{figure*}[!htb]
    \centering
    \includegraphics[width=0.95\linewidth]{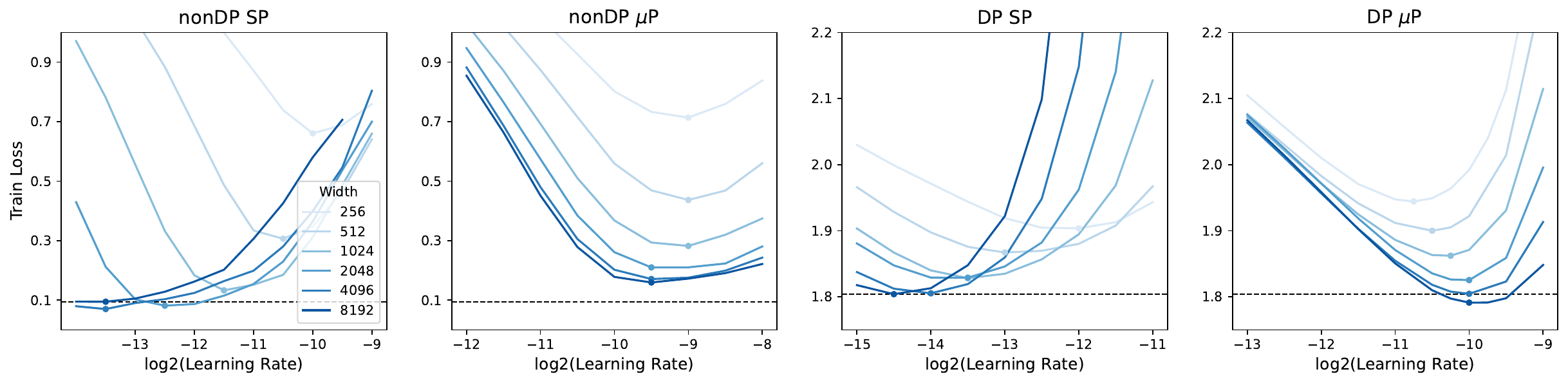}
    \caption{\textbf{Comparison between SP and $\mu$P (Adam).} Similar to SGD, we adjust the clipping coefficient and noise to satisfy the SNR condition. }
    \label{fig:mlp-adam-loss}
\end{figure*}

In this section, we provide the supplementary MLP experimental results referenced in the main text.
Figure~\ref{fig:mlp-adam-loss} illustrates the training loss of a 5-layer MLP on CIFAR-10 over 10 epochs across varying model widths.
We compare SP and $\mu$P with Adam under both non-private and differentially private settings.

\newpage
\subsection{Batch Size Transfer}
\label{app:batch-size}
\begin{figure*}[!htb]
    \centering
    \includegraphics[width=0.7125\linewidth]{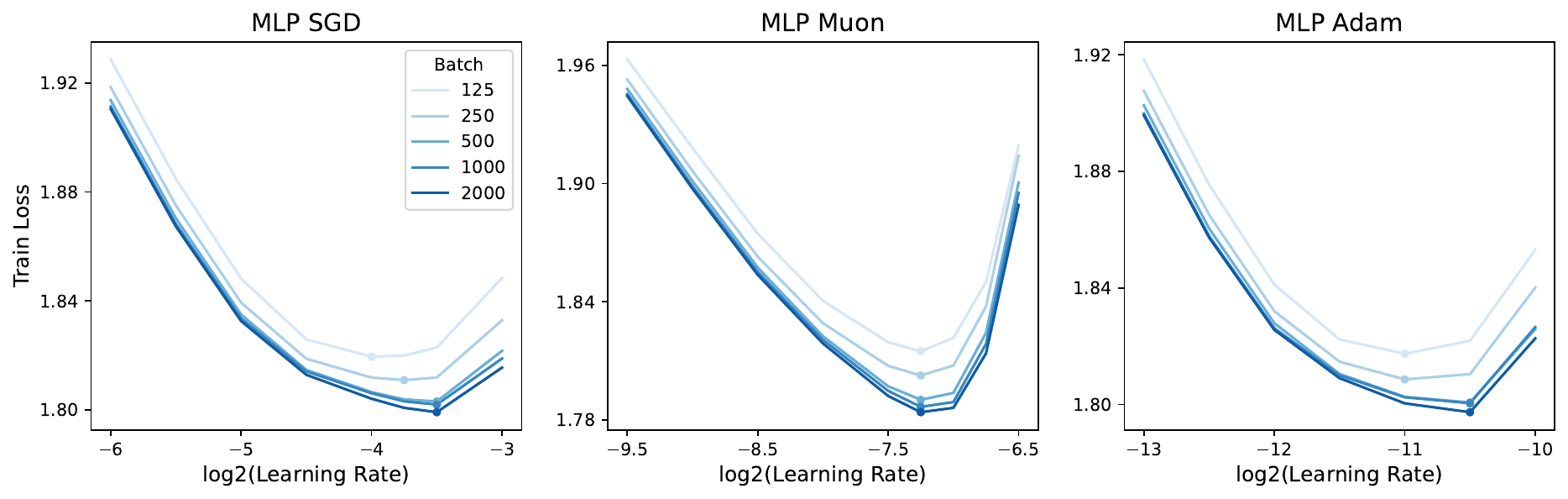}
    \caption{\textbf{Batch size transfer of 5-layer MLP on CIFAR-10.} We compare training loss across different batch sizes using SGD, Adam, and Muon optimizers. Curves align when plotted against iterations, confirming effective transfer.}
    \label{fig:batch-size-mlp}
\end{figure*}

\begin{figure*}[!htp]
    \centering
    \includegraphics[width=0.7125\linewidth]{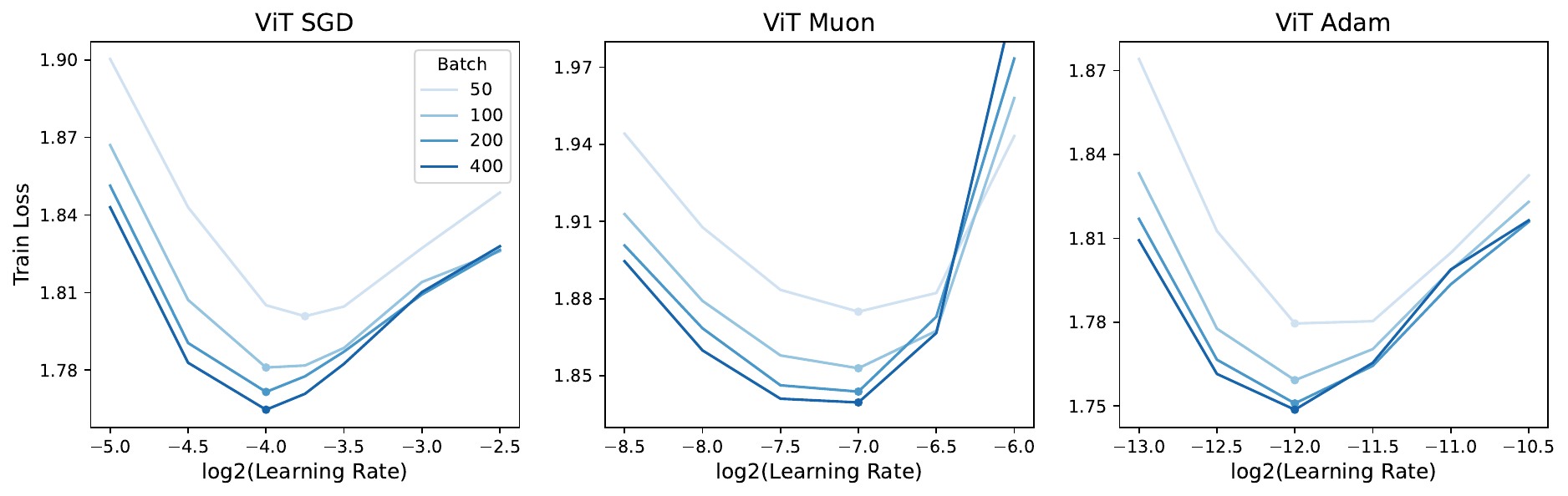}
    \caption{\textbf{Batch size transfer of ViT on CIFAR-10.} Similar to the MLP results, the ViT loss curves align across different batch sizes when plotted against iterations, confirming effective transfer for the Transformer architecture.}
    \label{fig:batch-size-vit}
\end{figure*}

In this section, we supplement our main analysis with batch size transfer experiments, as illustrated in Figures~\ref{fig:batch-size-mlp} and \ref{fig:batch-size-vit}. For a fixed model width, we adjust the noise multiplier $\sigma$ proportionally to the batch size. Crucially, to ensure a valid comparison of optimization trajectories, we fix the total number of training iterations across settings rather than the number of epochs.

{The noise multiplier and the batch size enter the noise-to-signal ratio only through $\sigma/B$, so scaling them together leaves the SNR unchanged.
Fixing the number of iterations also keeps the privacy budget approximately fixed as the batch size varies, since the sampling rate grows with $B$ at the same rate as $\sigma$.
Hence the transfer applies to any batch size, once it is paired with the noise multiplier returned by the privacy accountant.}

\subsection{ViT Supplementary Results}
\label{app:vit-sp}
\begin{figure*}[ht]
    \centering
    \includegraphics[width=0.95\linewidth]{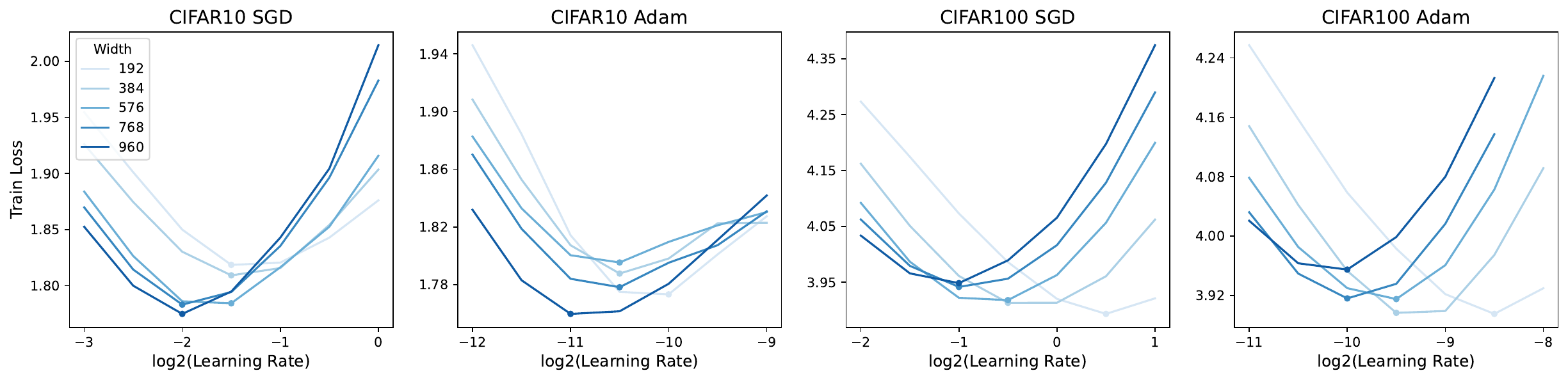}
    \caption{\textbf{Failure of Standard Parametrization (SP) on ViT.} As model width increases, the training dynamics under SP become unstable or diverge, failing to exhibit the hyperparameter transfer seen in $\mu$P.}
    \label{fig:vit-sp-failure}
\end{figure*}

\begin{figure*}[!htb]
    \centering
    \includegraphics[width=0.7125\linewidth]{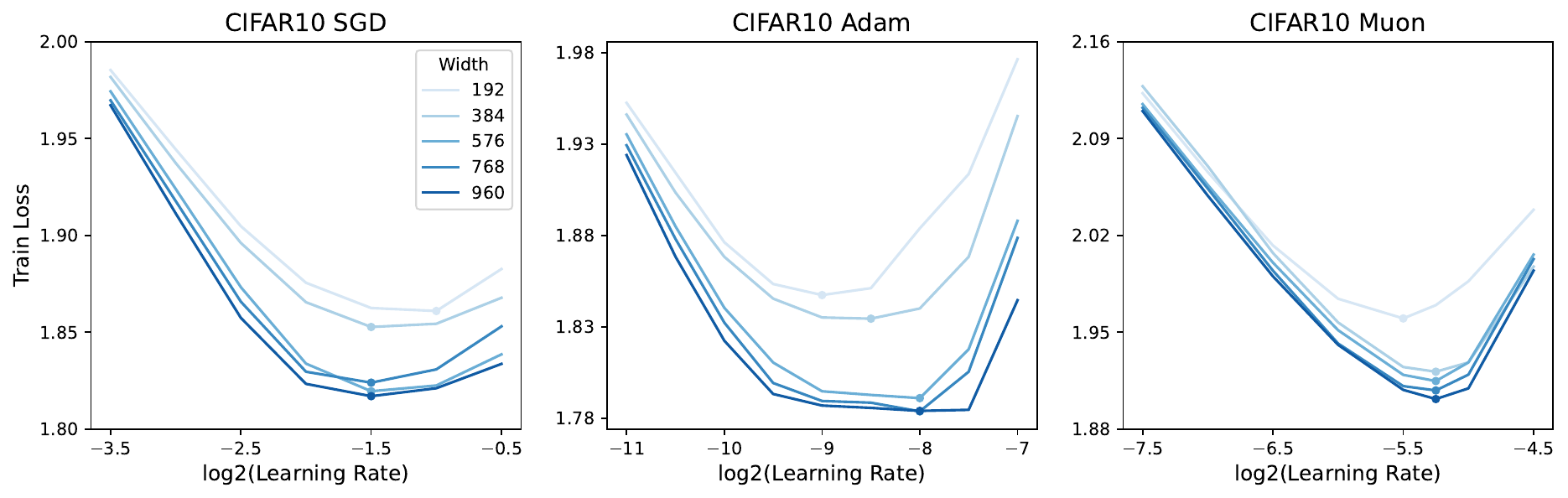}
    \caption{\textbf{$\mu$P ViT on CIFAR-10.} The optimizer configuration is SGD, Adam and Muon (with SGD boundary updates). }
    \label{fig:vit-cifar10}
\end{figure*}

In this section, we present supplementary evidence demonstrating the failure of Standard Parametrization (SP), alongside the results on CIFAR-10 for Vision Transformers (ViT). As shown in Figure \ref{fig:vit-sp-failure}, when trained with DP-SGD and DP-Adam, SP models fail to maintain consistent training dynamics across varying widths. In contrast, our DP-$\mu$P approach (Figure \ref{fig:vit-cifar10}) demonstrates perfect hyperparameter transfer.

\subsection{Final Utility and Robustness across Seeds}
\label{app:utility}
{We examine whether transfer also holds for the final test accuracy, and whether it is robust to the randomness of the DP noise and the initialization.
For the 5-layer MLP and the ViT on CIFAR-10, we grid search the learning rate at every width, select the one achieving the best test accuracy, and repeat each run over 5 seeds fixed in advance.
All runs use $(\epsilon, \delta) = (2, 10^{-5})$ for the target model, so the comparison is made at matched privacy.
As shown in Table~\ref{tab:utility-mlp}, the optimal learning rate under SP drifts with width, while under $\mu$P it stays nearly constant.
At the largest width, $\mu$P reaches slightly higher test accuracy than SP tuned directly at that width, for both SGD and Adam.
Table~\ref{tab:utility-vit} shows the same stability for the ViT, and in both tables the optimal learning rate is stable across seeds.
These optimal learning rates differ from those in the training-loss figures, since the learning rate minimizing training loss need not maximize test accuracy.}

\begin{table}[!htb]
  \caption{Optimal $\log_2$ learning rate (by test accuracy) at each width for a 5-layer MLP on CIFAR-10, reported as mean $\pm$ std over 5 seeds. Slope $\beta$ is the regression slope of the optimal $\log_2$ learning rate against $\log_2$ width, and Acc is the test accuracy (\%) at the largest width.}
  \label{tab:utility-mlp}
  \centering
  \small
  \setlength{\tabcolsep}{4pt}
  \renewcommand{\arraystretch}{1.15}
  \begin{tabular}{r|cc|cc|c}
    \toprule
    width & SP SGD & $\mu$P SGD & SP Adam & $\mu$P Adam & $\mu$P Muon \\
    \midrule
    256  & $-1.50\pm0.25$ & $-3.55\pm0.21$ & $-10.50\pm0.18$ & $-9.75\pm0.18$ & $-8.25\pm0.18$ \\
    512  & $-1.70\pm0.27$ & $-3.60\pm0.14$ & $-10.90\pm0.29$ & $-9.65\pm0.22$ & $-8.75\pm0.31$ \\
    1024 & $-2.25\pm0.25$ & $-3.25\pm0.00$ & $-11.10\pm0.22$ & $-9.75\pm0.18$ & $-8.55\pm0.11$ \\
    2048 & $-2.85\pm0.14$ & $-3.15\pm0.14$ & $-11.55\pm0.21$ & $-9.15\pm0.14$ & $-8.75\pm0.25$ \\
    4096 & $-3.45\pm0.21$ & $-2.90\pm0.14$ & $-12.30\pm0.21$ & $-9.10\pm0.14$ & $-8.65\pm0.22$ \\
    8192 & $-4.00\pm0.18$ & $-2.90\pm0.14$ & $-12.85\pm0.14$ & $-8.90\pm0.14$ & $-8.65\pm0.14$ \\
    \midrule
    Slope $\beta$ & $-0.52$ & $+0.16$ & $-0.47$ & $+0.19$ & $-0.05$ \\
    Acc & $41.16\pm0.23$ & $41.77\pm0.38$ & $41.14\pm0.29$ & $41.80\pm0.28$ & $40.76\pm0.16$ \\
    \bottomrule
  \end{tabular}
\end{table}

\begin{table}[!htb]
  \caption{Optimal $\log_2$ learning rate (by test accuracy) at each width for a ViT on CIFAR-10, reported as mean $\pm$ std over 5 seeds. Slope $\beta$ and Acc are defined as in Table~\ref{tab:utility-mlp}.}
  \label{tab:utility-vit}
  \centering
  \small
  \setlength{\tabcolsep}{4pt}
  \renewcommand{\arraystretch}{1.15}
  \begin{tabular}{r|ccc}
    \toprule
    width & $\mu$P SGD & $\mu$P Adam & $\mu$P Muon \\
    \midrule
    192 & $-1.35\pm0.34$ & $-8.50\pm0.25$ & $-5.85\pm0.14$ \\
    384 & $-1.40\pm0.22$ & $-8.05\pm0.11$ & $-5.70\pm0.11$ \\
    576 & $-1.00\pm0.35$ & $-8.10\pm0.29$ & $-5.65\pm0.14$ \\
    768 & $-0.95\pm0.45$ & $-8.10\pm0.22$ & $-5.80\pm0.21$ \\
    960 & $-1.15\pm0.22$ & $-8.30\pm0.21$ & $-5.65\pm0.14$ \\
    \midrule
    Slope $\beta$ & $+0.16$ & $+0.10$ & $+0.06$ \\
    Acc & $33.37\pm0.43$ & $36.79\pm0.21$ & $30.08\pm0.41$ \\
    \bottomrule
  \end{tabular}
\end{table}

\subsection{Standard Parametrization Coordinate Check}
\label{app:coord-check-sp}
\begin{figure*}[ht]
    \centering
    \includegraphics[width=0.95\linewidth]{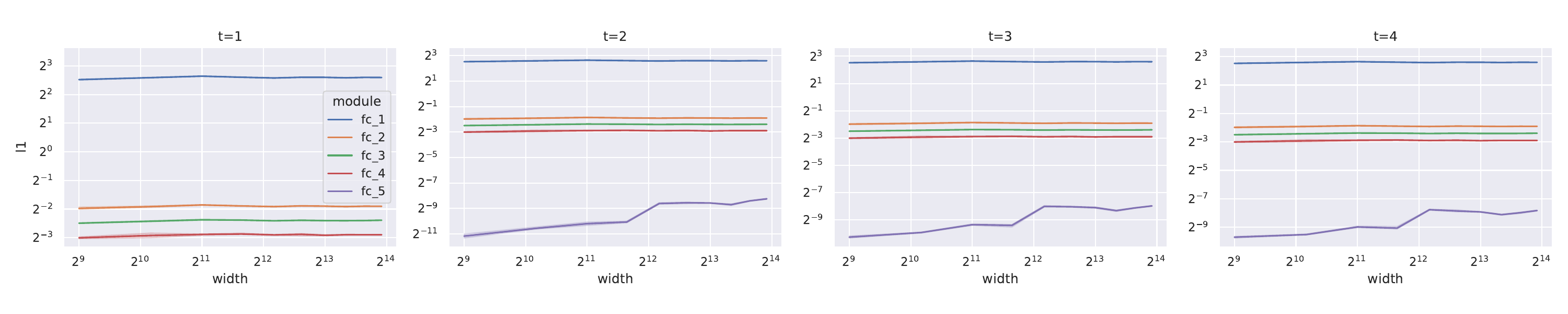}
    \caption{\textbf{Coordinate check for MLP (SP) with SNR condition.} Even with the SNR condition applied, standard parametrization often fails to preserve feature scale across widths.}
    \label{fig:mlp-sp-snr}
\end{figure*}

\begin{figure*}[!htb]
    \centering
    \includegraphics[width=0.95\linewidth]{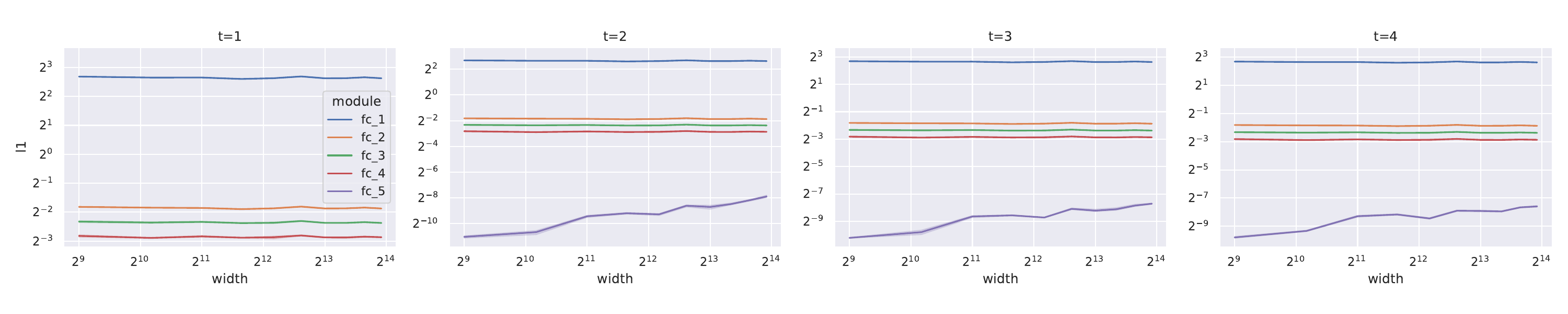}
    \caption{\textbf{Coordinate check for MLP (SP) without SNR condition.} Without privacy-aware scaling, feature activations diverge significantly as width increases.}
    \label{fig:mlp-sp-no-snr}
\end{figure*}

\begin{figure*}[!htb]
    \centering
    \includegraphics[width=0.95\linewidth]{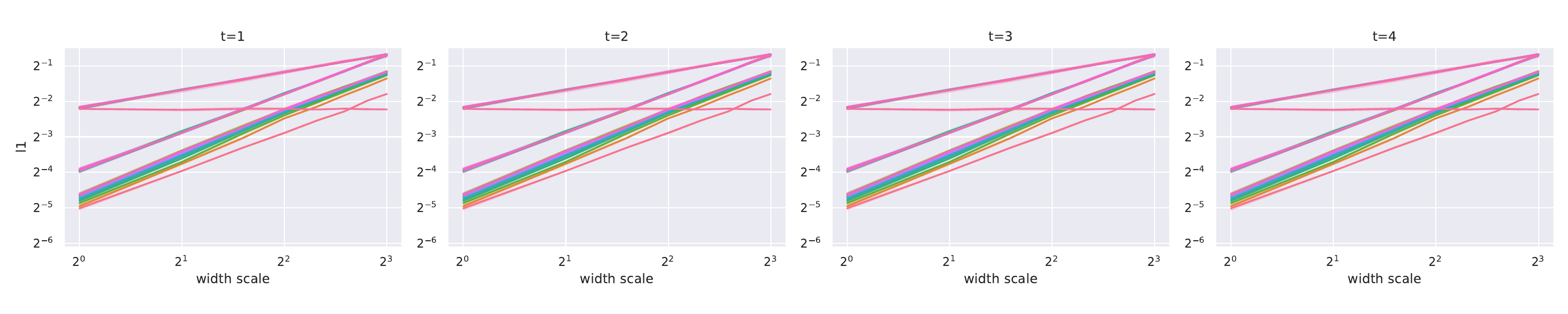}
    \caption{\textbf{Coordinate check for ViT (SP).} Similar to the MLP case, SP leads to inconsistent activation scales in Vision Transformers.}
    \label{fig:vit-sp-coord}
\end{figure*}

We visualize the coordinate-wise activation norms for Standard Parametrization in Figures~\ref{fig:mlp-sp-snr}, \ref{fig:mlp-sp-no-snr} and \ref{fig:vit-sp-coord}, to contrast with the stable behavior of $\mu$P shown in the main text.

\subsection{Illustrative Example of Parameter Derivation}
\label{sec:derivation_example}
In this section, we provide a concrete parameter derivation to illustrate the DP-$\mu$P framework depicted in Figure~\ref{fig:flowchart}. We consider a 5-layer MLP trained on CIFAR-10, where the input dimension is $n_0 = 3072$ ($3 \times 32 \times 32$) and the output dimension is $n_5 = 10$. Our objective is to transfer hyperparameters from a small proxy model with hidden widths $n_{1,\dots,4}^\prime = 128$ to a large target model with widths $n_{1,\dots,4} = 8192$.

{Before the search, we fix the proxy width, the target width, the privacy budget, the batch size and the number of training steps, since the last three determine the noise multiplier.}

\paragraph{Step 1: Calibrate the noise multiplier of the target.}
For a target privacy budget of $(\epsilon, \delta) = (2, 10^{-5})$ on the CIFAR-10 dataset ($50,000$ samples), we configure the training with a batch size of $B=500$ and train for $10$ epochs. Based on these parameters, the privacy accountant computes the required noise multiplier for the large model to be $\sigma \approx 1.0229$, which we approximate as $\sigma \approx 1$.

\paragraph{Step 2: Adopt uniform clipping on the proxy.}
To facilitate the hyperparameter search on the proxy model, we employ a uniform clipping (e.g. $D_l^\prime = 5$). We use uniform clipping on the proxy in all experiments.

\paragraph{Step 3: Solve the SNR constraints.}
By solving the SNR constraints defined in \cref{eq:scaling_squared}, we obtain the required noise multiplier for the proxy model:
\begin{equation}
    \sigma^\prime \approx 6.91.
\end{equation}
Simultaneously, we derive the layer-wise scaling coefficients $\frac{1}{\sqrt{\bm{D}}}$ for the target large model:
\begin{equation}
    \bm{D} \approx (49.9, 3.73, 3.73, 3.73, 5.7) \Longrightarrow \frac{1}{\sqrt{\bm{D}}} \approx (0.14, 0.52, 0.52, 0.52, 0.42).
\end{equation}
This specific configuration ensures that the Signal-to-Noise Ratio (SNR) is preserved across widths.

\paragraph{Step 4: Search on the proxy.}
We proceed to perform hyperparameter tuning on the small proxy model using the derived noise multiplier $\sigma^\prime \approx 6.91$ and a uniform clipping coefficient $1/\sqrt{D^\prime} = 1/\sqrt{5}$. We conduct a grid search to identify the optimal learning rate. For illustrative simplicity, we assume this search yields an optimal value of $\eta_{\text{small}}^* = 1$. Note that we employ a global learning rate for the small model, as the primary objective at this stage is to ensure numerical stability (preventing gradient vanishing or explosion) rather than strictly enforcing layer-wise alignment.

\paragraph{Step 5: Transfer to the target.}
To determine the layer-wise learning rates for the large target model, we scale the optimal proxy rate $\eta_{\text{small}}^* = 1$ based on the spectral norm ratios, in accordance with Lemma~\ref{lem:layerwise lr}. The transfer rule is given by:
\begin{equation}
\label{eq:lr_transfer}
    \frac{\eta_{\text{large}}^{(l)}}{\eta_{\text{small}}^*} = \frac{\sqrt{n_l/n_{l-1}}/\left(\sigma\cdot\left(\sqrt{n_l} + \sqrt{n_{l-1}}\right)\right)}{\sqrt{n_l^\prime/n_{l-1}^\prime}/\left(\sigma^\prime\cdot\left(\sqrt{n_l^\prime} + \sqrt{n_{l-1}^\prime}\right)\right)},
\end{equation}
which results in $\bm{\eta}_{\text{large}} \approx (25.28,\; 0.86,\; 0.86,\; 0.86,\; 0.13)$.
The rest of the training recipe found on the proxy is reused on the target as it is.

\paragraph{Selecting the batch size.}
If the batch size is also to be selected, the noise multiplier has to be recomputed for each candidate batch size.
Steps 1--5 are then repeated for each candidate, and the best candidate is chosen on the proxy model.

\subsection{End-to-End Privacy Accounting}
\label{sec:end_to_end_privacy}

We note that DP-$\mu$P serves as a method to transferring hyperparameters for DP optimization without privacy cost. We distinguish the privacy cost of hyperparameter search from that of
training the final target model, providing an end-to-end privacy accounting. 

Let $\mathcal{D}$ denote the private dataset.
We first construct the proxy training configuration according to the DP-$\mu$P scaling recipe: the target privacy parameters and the two architectures determine, through the SNR-matching constraints in \eqref{eq:scaling_squared}, the proxy noise multiplier and the layer-wise clipping parameters of the target.
This step depends only on the privacy parameters and the architectures and does not access $\mathcal{D}$, so it incurs no privacy cost.

Hyperparameter search is then performed only over proxy runs generated
under this DP-$\mu$P-aligned configuration.
The selected proxy learning rate is subsequently
mapped to the target-model layer-wise learning rates using the deterministic transfer {rule $T$, given by Lemma~\ref{lem:layerwise lr} and illustrated in Appendix~\ref{sec:derivation_example}.}

\paragraph{Privacy cost of hyperparameter search.}
Suppose the complete search-and-selection mechanism
$\mathcal{M}_{\mathrm{search}}$ is
$(\epsilon_{\mathrm{search}},\delta_{\mathrm{search}})$-DP.
Importantly, this privacy guarantee must account for the entire search
procedure, rather than a single proxy-model training run. For example,
if multiple candidate learning rates are evaluated on the same private
dataset, the privacy costs of those evaluations and the subsequent
selection must be incorporated into
$(\epsilon_{\mathrm{search}},\delta_{\mathrm{search}})$.
Alternatively, the search may be performed using public data, in which
case
$\epsilon_{\mathrm{search}}=\delta_{\mathrm{search}}=0$.

\paragraph{Privacy cost of target-model training.}
For every fixed transferred hyperparameter configuration
$\hat{\eta}_{\mathrm{target}}$, suppose
$\mathcal{M}_{\mathrm{train}}
(\mathcal{D};\hat{\eta}_{\mathrm{target}})$ is
$(\epsilon_{\mathrm{train}},\delta_{\mathrm{train}})$-DP.
Although the target-model training procedure depends adaptively on the
private output of $\mathcal{M}_{\mathrm{search}}$, adaptive sequential composition shows that the end-to-end mechanism
\[
    \mathcal{M}_{\mathrm{train}}
    \!\left(
        \mathcal{D};
        T(\mathcal{M}_{\mathrm{search}}(\mathcal{D}))
    \right)
\]
satisfies
\[
    \left(
        \epsilon_{\mathrm{search}}+\epsilon_{\mathrm{train}},
        \;
        \delta_{\mathrm{search}}+\delta_{\mathrm{train}}
    \right)\text{-DP}.
\]
Hence, for a prescribed total privacy budget
{$(\epsilon_{\mathrm{total}},\delta_{\mathrm{total}})$}, it is sufficient to allocate the budgets such that
\[
    \epsilon_{\mathrm{search}}
    +
    \epsilon_{\mathrm{train}}
    \leq \epsilon_{\mathrm{total}},
    \qquad
    \delta_{\mathrm{search}}
    +
    \delta_{\mathrm{train}}
    \leq \delta_{\mathrm{total}}.
\]
A simple parameterization is
\[
    \epsilon_{\mathrm{search}}=\gamma\epsilon_{\mathrm{total}},
    \qquad
    \epsilon_{\mathrm{train}}=(1-\gamma)\epsilon_{\mathrm{total}},
\]
and
\[
    \delta_{\mathrm{search}}=\gamma_\delta\delta_{\mathrm{total}},
    \qquad
    \delta_{\mathrm{train}}=(1-\gamma_\delta)\delta_{\mathrm{total}},
\]
for predetermined
$\gamma,\gamma_\delta\in[0,1]$ before accessing $\mathcal{D}$.

\paragraph{Tighter accounting.}
The additive guarantee above follows directly from basic sequential
composition and provides the simplest end-to-end accounting rule. In
practice, tighter guarantees can be obtained by carrying out both
stages in a common privacy accountant. For example, under R\'enyi
differential privacy, if the two stages satisfy
$\rho_{\mathrm{search}}(\alpha)$-RDP and
$\rho_{\mathrm{train}}(\alpha)$-RDP at order $\alpha$, respectively,
then
\[
    \rho_{\mathrm{total}}(\alpha)
    =
    \rho_{\mathrm{search}}(\alpha)
    +
    \rho_{\mathrm{train}}(\alpha),
\]
and the composed guarantee can be converted once to
$(\epsilon_{\mathrm{total}},\delta_{\mathrm{total}})$-DP at the end. More refined hyperparameter-tuning
or privacy-loss-distribution accountants (such as GDP) may likewise be used without
changing our transfer mechanism.


\end{document}